\documentclass[conference]{IEEEtran}
\IEEEoverridecommandlockouts
\usepackage{cite}
\usepackage{amsmath,amssymb,amsfonts}
\usepackage[ruled,vlined,linesnumbered,noresetcount]{algorithm2e}
\usepackage{algorithmic}
\usepackage{graphicx}
\usepackage{textcomp}
\usepackage{xcolor}
\usepackage{bm}  % Include the bm package for bold symbols
\SetAlgoNoEnd % This will suppress the "end" keywords
\usepackage{caption}
\usepackage{subcaption}
\usepackage{multirow}
\usepackage{amsmath}

\newtheorem{theorem}{Definition}
\newtheorem{condition}{Condition}
\usepackage{colortbl}
\SetKwInput{KwInput}{Input}                % Set the Input
\SetKwInput{KwOutput}{Output}

\def\BibTeX{{\rm B\kern-.05em{\sc i\kern-.025em b}\kern-.08em
    T\kern-.1667em\lower.7ex\hbox{E}\kern-.125emX}}
\begin{document}

\title{Tackling Oversmoothing in GNN via Graph Sparsification: A Truss-based Approach }

% \author{\IEEEauthorblockN{Anonymous Authors}}
% 

\author{
    \IEEEauthorblockN{Tanvir Hossain\IEEEauthorrefmark{1},
    Khaled Mohammed Saifuddin\IEEEauthorrefmark{1},
    Muhammad Ifte Khairul Islam\IEEEauthorrefmark{1},
    Farhan Tanvir\IEEEauthorrefmark{1},
    Esra Akbas\IEEEauthorrefmark{1}}
    \IEEEauthorblockA{\IEEEauthorrefmark{1}%
    Department of Computer Science, Georgia State University, Atlanta, GA 30302, USA\\
    \{thossain5, ksaifuddin1, mislam29\}@student.gsu.edu, \{ftanvir, eakbas1\}@gsu.edu}
}

\maketitle
\newcommand{\TGS}{\texttt{TGS}}
\begin{abstract}
Graph Neural Network (GNN) achieves great success for node-level and graph-level tasks via encoding meaningful topological structures of networks in various domains, ranging from social to biological networks. However, repeated aggregation operations lead to excessive mixing of node representations, particularly in dense regions with multiple GNN layers, resulting in nearly indistinguishable embeddings. This phenomenon leads to the oversmoothing problem that hampers downstream graph analytics tasks. To overcome this issue, we propose a novel and flexible truss-based graph sparsification model that prunes edges from dense regions of the graph. Pruning redundant edges in dense regions helps to prevent the aggregation of excessive neighborhood information during hierarchical message passing and pooling in GNN models. We then utilize our sparsification model in the state-of-the-art baseline GNNs and pooling models, such as GIN, SAGPool, GMT, DiffPool, MinCutPool, HGP-SL, DMonPool, and AdamGNN. Extensive experiments on different real-world datasets show that our model significantly improves the performance of the baseline GNN models in the graph classification task.

\end{abstract}


\section{Introduction}
In recent years, graph neural networks (GNN) have given promising performance in numerous applications over different domains, such as gene expression analysis~\cite{saifuddin2023hygnn}, traffic flow forecasting~\cite{zhang2021graph}, fraud detection~\cite{van2022inductive}, and recommendation system~\cite{deng2022recommender}. GNN effectively learns the representation of nodes and graphs via encoding topological graph structures into low-dimensional space through message passing and aggregation mechanisms. To learn the higher-order relations between nodes, especially for large graphs, we need to increase the number of layers. However, creating an expressive GNN model by adding more convolution layers increases redundant receptive fields for computational nodes and results in oversmoothing as node representations become nearly indistinguishable.

Several research works illustrate that due to oversmoothing, nodes lose their unique characteristics~\cite{huang2023hub,chen2020measuring}, adversely affecting GNNs' performance on downstream tasks, including node and graph classification. Different models have been proposed to overcome the problem, such as skip connection~\cite{chen2020simple}, drop edge~\cite{rong2019dropedge}, GraphCON~\cite{rusch2022graph}. While many of these methods focus on node classification, they often overlook the impact of oversmoothing on the entire network's representation. Additionally, only a limited number of studies have investigated the influence of specific regions causing oversmoothing ~\cite{chen2020simple,huang2023hub} in GNNs. These studies show that the smoothness in GNN varies for complex connections in different graph areas, and an individual node with high degrees converges to stationary states earlier than lower-degree nodes. Hence, the networks' regional structures affect the phenomenon because repeated message passing occurs within the dense neighborhood regions of the nodes. Therefore, we observe the impact of congested graph regions on oversmoothing.

\begin{figure}[b!]
    \centering
    % First row of subfigures
    \begin{subfigure}{0.25\textwidth}
        \centering
        \includegraphics[width=0.9\textwidth]{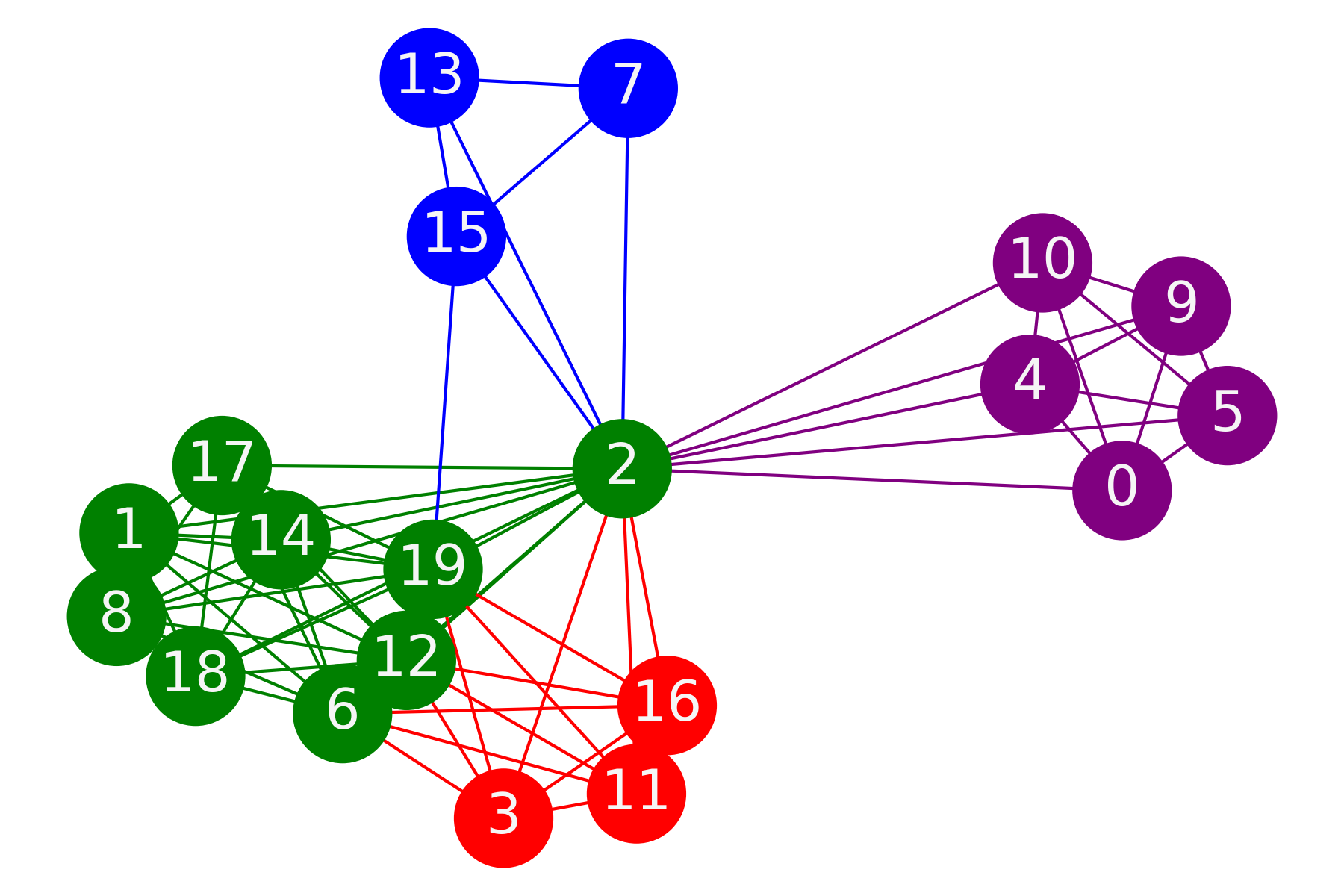}
        \caption{Experiment Graph}
        \label{fig: example_graph}
    \end{subfigure}%
    ~
    \begin{subfigure}{0.25\textwidth}
        \centering
        \includegraphics[width=0.95\textwidth]{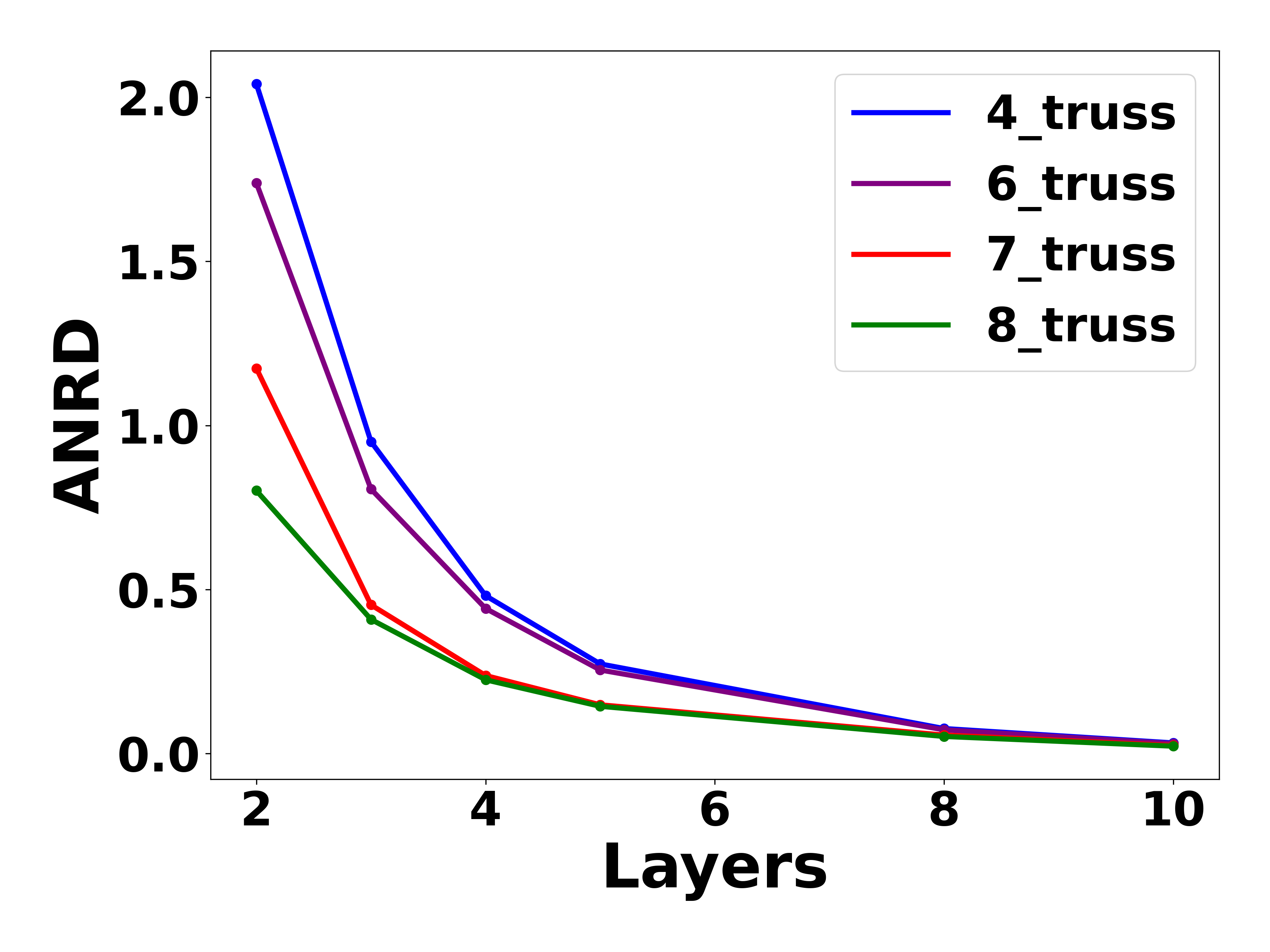}
        \caption{ANRD $vs$ Layers}
        \label{fig: ANRD_at_layers}
    \end{subfigure}

    \caption{Early Oversmoothing: Average node representation distance(ANRD) of different $k$-truss subgraphs concerning GNN layers (2-10)}
    \label{fig: early_oversmoothing}
\end{figure}
We conduct a small experiment to demonstrate the early oversmoothing at highly connected regions on a toy graph (Figure~\ref{fig: example_graph}). To calculate the density on the graph, we utilize the $k$-truss~\cite{akbas2017truss}, one of the widely used cohesive subgraph extraction models based on the number of triangles each edge contains. To show the smoothness of the node features, we utilize the average node representation distance (ANRD)~\cite{kipf2016semi}. We measure the ANRD of different $k$-truss regions and present how it changes through the increasing number of layers in GNN. We present the toy graph and ANRD values with respect to the number of layers in Figure~\ref{fig: ANRD_at_layers}. While the toy graph in the figure is a 4-truss graph, it has 6, 7, and 8-truss subgraphs. Nodes and edges are colored based on their trussness. As known, $k$-truss subgraphs have hierarchical relations, e.g., 7 and 8-truss subgraphs are included in the 6-truss subgraph. Even at layer 2, we observe the ANRD of 7 and 8-truss subgraphs substantially degrades compared to the lower truss $(k = 4, 6)$ subgraphs.

While oversmoothing is observed at the node level, it may also result in losing crucial information for the graphs' representation to distinguish them. Furthermore, to learn the graph representation, GNNs employ various hierarchical pooling approaches, including hierarchical coarsening and message-passing, resulting in oversmoothing via losing unique node features~\cite{wang2021exploring,zuo2023exploring}. Consequently, dense regions' identical node information affects the graph's representation learning.

\begin{figure*}[t]
    \centering
    \begin{subfigure}[h]{0.24\textwidth}
        \centering
        \includegraphics[width=\textwidth]{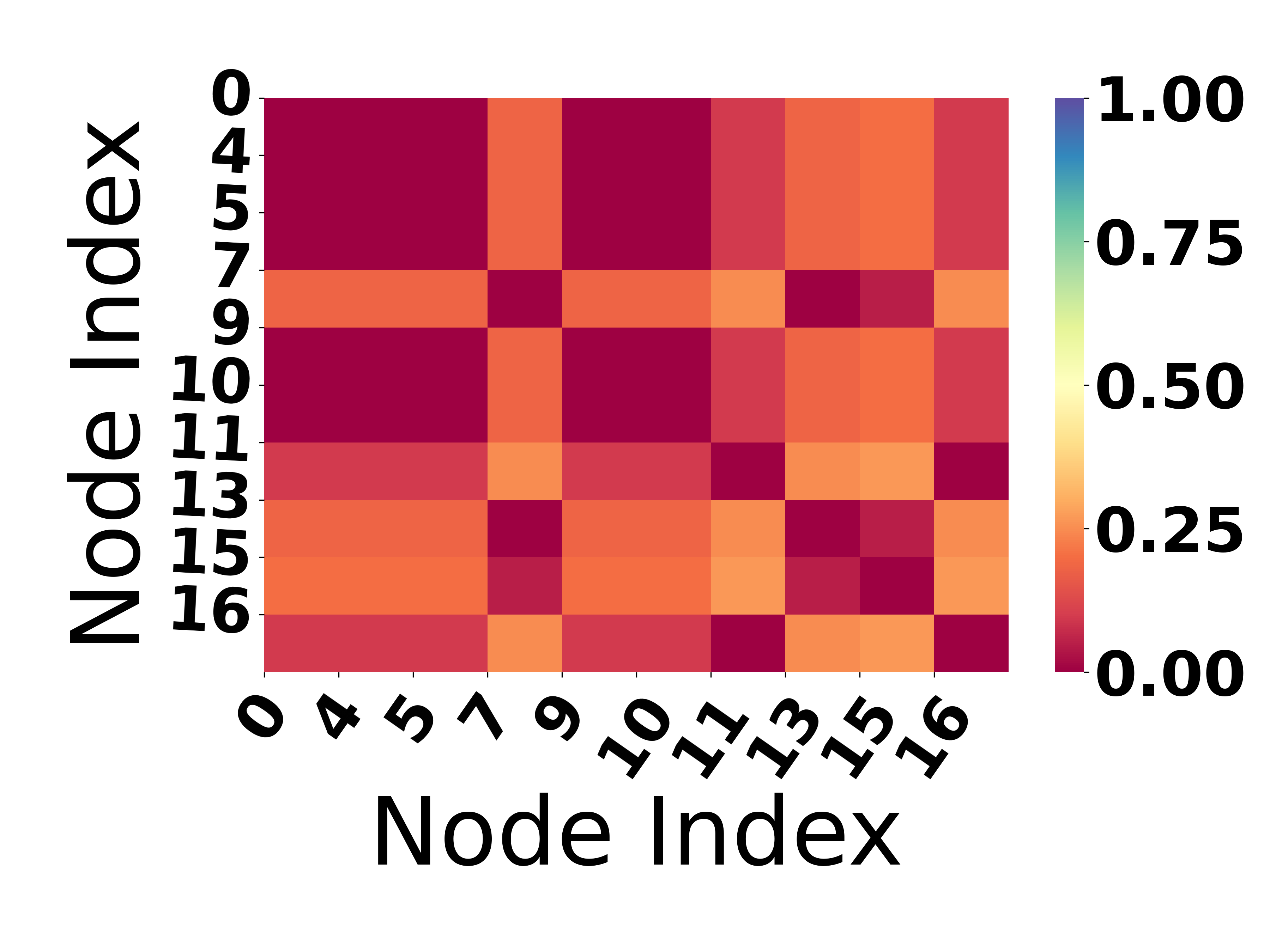}
        \caption{Org. pool 1}
        \label{fig: pool1_org}
    \end{subfigure}%
     ~
    \centering
    \begin{subfigure}[h]{0.24\textwidth}
        \centering
        \includegraphics[width=\textwidth]{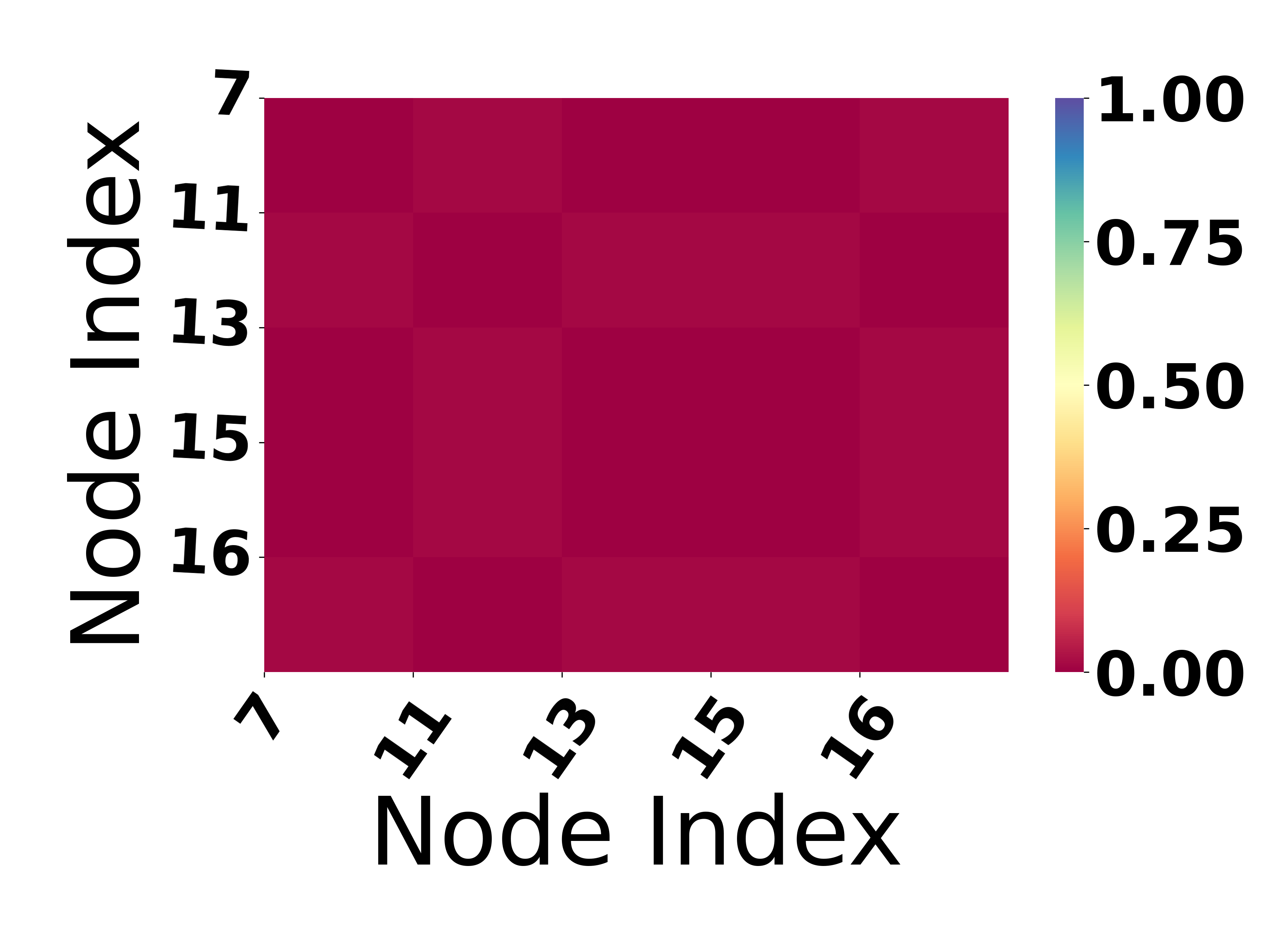}
        \caption{Org. pool 2}
        \label{fig: pool2_org}
    \end{subfigure}%
    ~
    \begin{subfigure}[h]{0.24\textwidth}
        \centering
        \includegraphics[width=\textwidth]{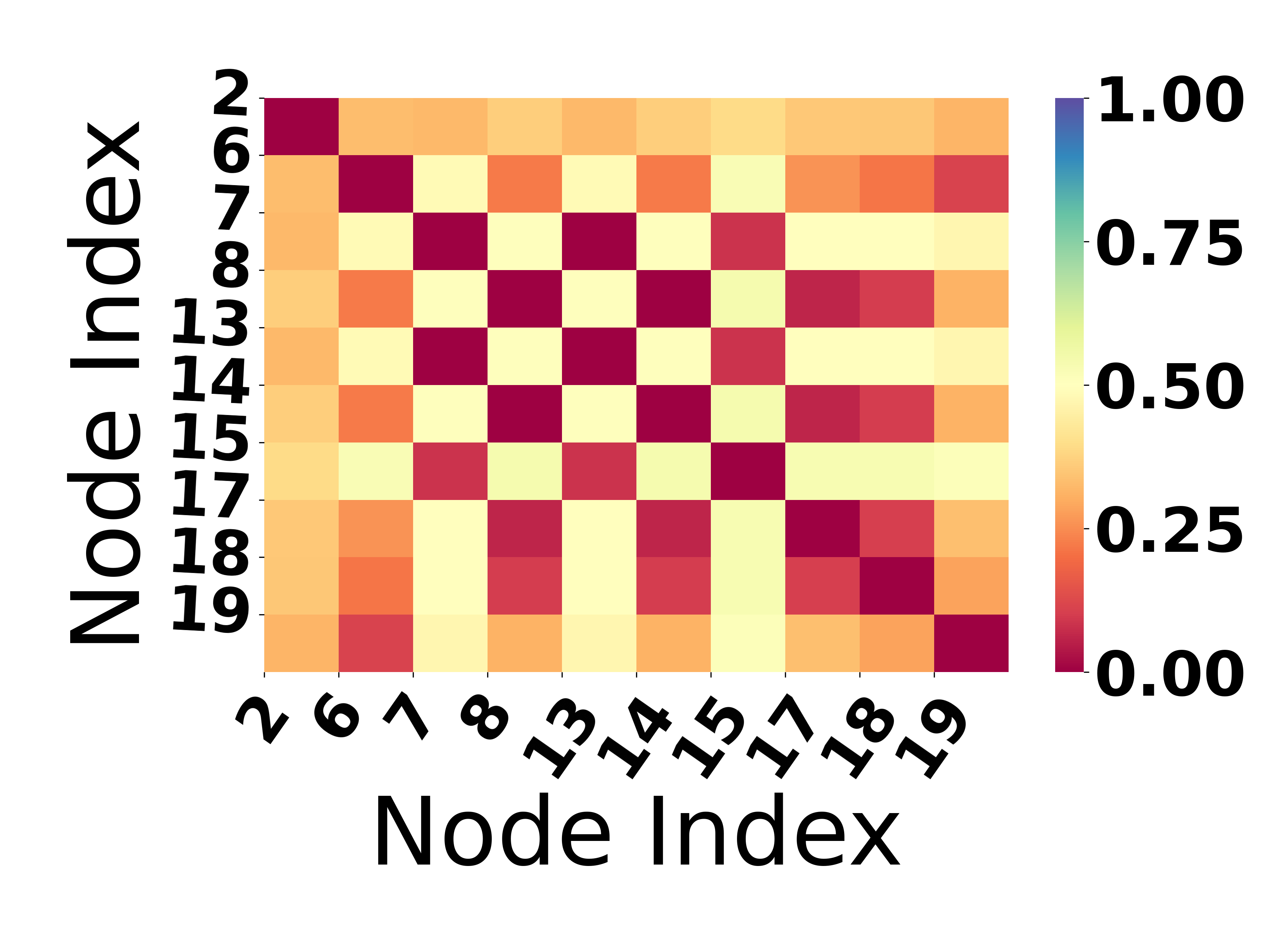}
        \caption{Spars. pool 1}
        \label{fig:spars_pool1}
    \end{subfigure}%
    ~
    \begin{subfigure}[h]{0.24\textwidth}
        \centering
        \includegraphics[width=\textwidth]{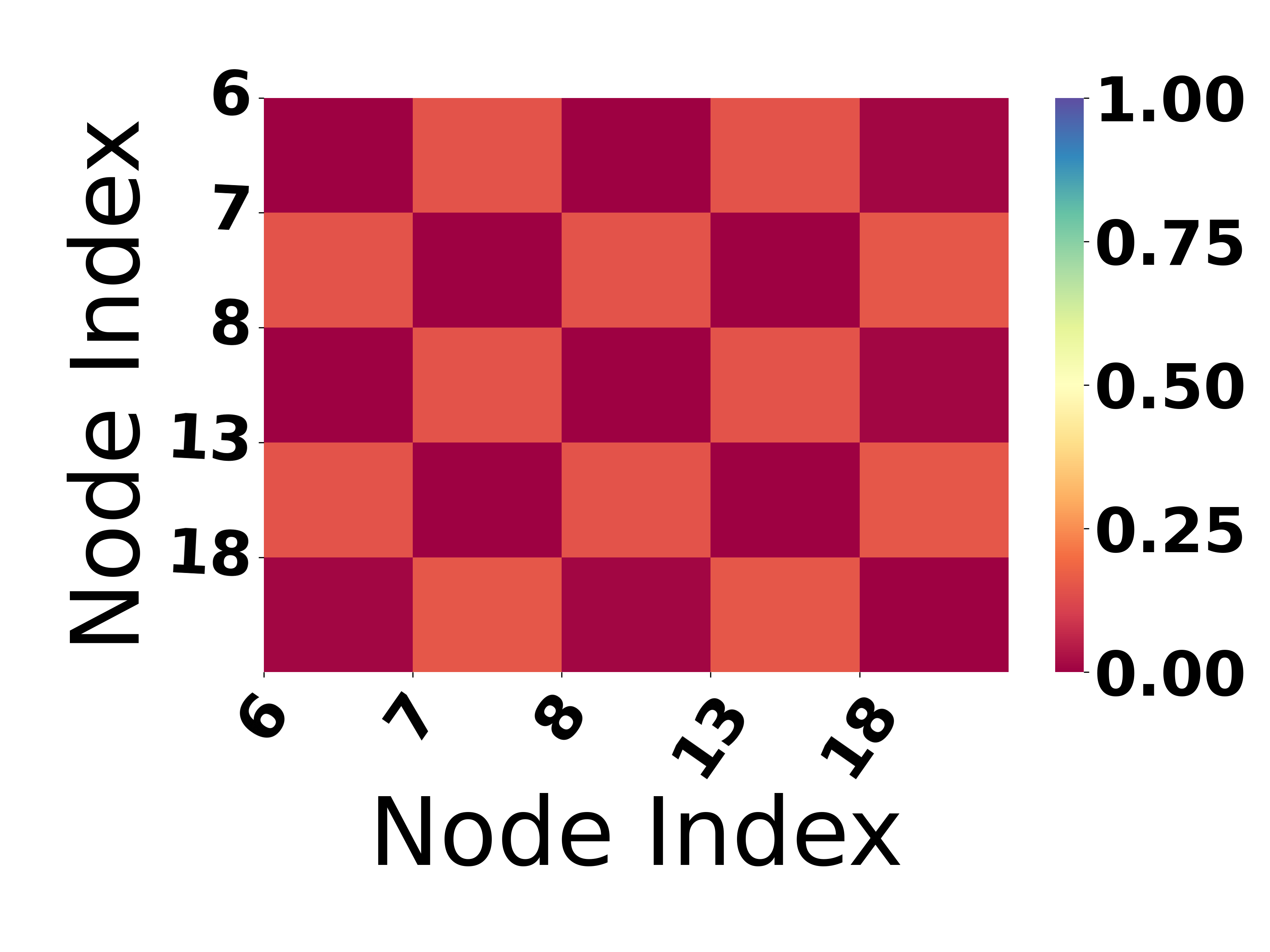}
        \caption{Spars. pool 2}
        \label{fig:spars_pool2}
    \end{subfigure}%
    \caption{ESM of the toy graph and sparsified graph for $\delta = 7$}
    \label{fig: pruned_emb_distance}
\end{figure*}

We extend the preliminary investigation on the toy graph given in Figure~\ref{fig: example_graph}. We first apply the SAGPool model. After each pooling layer's operation, we measure the coarsened graph's nodes' embedding space matrix (ESM) with $l_2$ norm, then present the results for the first 2 pooling layers in Figure~\ref{fig: pool1_org} and \ref{fig: pool2_org}. We observe that embedding distances are getting smaller for nodes within the dense regions, significantly reducing the final graph's representation variability. These node and graph representation characteristics through GNN models inspire us to work at different levels of dense regions in the network to mitigate oversmoothing.

\textbf{Our Work}. To tackle the challenge, we develop a truss-based graph sparsification (\TGS) model. Earlier sparsification models apply supervised techniques~\cite{zheng2020robust} and randomly drop edges~\cite{rong2019dropedge,huang2023hub} which may result in losing meaningful connections. However, our model selects the initial extraneous information propagating edges by utilizing edge trussness. It operates on the candidate edge's nodes' neighborhood and measures the connectivity strength of nodes. This connectivity strength assists in understanding the edge's local and global impact on GNN's propagation steps. Based on their specific node strength limits, we decide which edges to prune and which to keep. Removing selected redundant edges from dense regions reduces noisy message passing. That decreases oversmoothing and facilitates the GNN's consistency in performance during training and inference time. As we see in Figure \ref{fig:spars_pool1} and \ref{fig:spars_pool2}, the sparsified graph exhibits greater diversity in node distances than the original, enhancing better representation learning. In a nutshell, the contributions of our model are listed as follows. 
\begin{itemize}
    \item We observe the prior stationary representation learning of nodes emerging in the network's high-truss region, which denotes a new perspective on explaining oversmoothing. We develop a unique truss-based graph sparsification technique to resolve this issue. 

    \item In the edge pruning step, we measure the two nodes' average neighborhood trussness to detect the regional interconnectedness strength of the nodes. During the message passing steps in GNN, as we trim down the dense connections within subgraphs, nodes in less dense areas at varying hop distances acquire diverse hierarchical neighbor information. Conversely, nodes in highly dense regions receive reduced redundant information. This provides smoothness to the node representation as well as to the graph representation. 

    \item We provide a simple but effective model by pruning noisy edges from graphs based on their nodes' average neighborhood trussness. The effectiveness of our model has been evaluated in comparison with standard GNN and graph pooling models. Extensive experiments on different real-world graphs show that our approach outperforms most of those baselines in graph classification tasks.
    
\end{itemize}

The rest of this paper is organized as follows. Section~\ref{sec:related_works} discusses the related work that informs our research. Section~\ref{sec:preliminaries} introduces the model's preliminaries, whereas Section~\ref{sec:methodology} describes the model itself. Next,  Section~\ref{sec: exp_and_res} represents our model's experiment results and an analysis of its performance on different datasets. Finally, we conclude the paper with a discussion of future research directions.

\section{Related Works}\label{sec:related_works}
\textbf{Graph Classification}. Early GNN models leverage simple readout functions to embed the entire graph. GIN~\cite{xu2018powerful} introduces their lack of expressivity and employs deep multiset sums to represent graphs. In recent years, graph pooling methods have acquired excellent traction for graph representation. They consider essential nodes' features instead of all nodes.
% These pooling models are divided into Flat pooling and Hierarchical pooling. 
Flat pooling methods utilize the nodes' representation without considering their hierarchical structure. Among them, GMT~\cite{baek2021accurate} proposes a multiset transformer for capturing nodes' hierarchical interactions. Another approach, SOPool~\cite{wang2020second}, capitalizes vertices second-order statistics for pooling graphs.

There are two main types of hierarchical pooling methods: clustering-based and selection-based. 
Clustering-based methods assign nodes to different clusters: computing a cluster assignment matrix~\cite{ying2018hierarchical}, utilizing modularity~\cite{tsitsulin2023graph} or spectral clustering~\cite{bianchi2020spectral} from the node's features and adjacency. On the other hand, selection-based models compute nodes' importance scores up to different hop neighbors and select essential nodes from them. Two notable methods are SAGPool~\cite{lee2019self}, which employs a self-attention mechanism to compute the node importance, and HGP-SL~\cite{zhang2019hierarchical}, which uses a sparse-max function to pool graphs. KPLEXPOOL~\cite{bacciu2021k} hierarchically leverages k-plex and graph covers to capture essential graph structures and facilitates the diffusion between contexts of distance nodes. Some approaches combine both hierarchical pooling types to represent graphs. One model, ASAP~\cite{ranjan2020asap}, adapted a new self-attention mechanism for node sectioning, a convolution variant for cluster assignment. Another model, AdamGNN~\cite{zhong2022multi}, employs multi-grained semantics for adapting selection and clustering for pooling graphs. 

\textbf{Oversmoothing:} While increasing number of layers for a regular neural network may results better learning, it may cause an oversmoothing problem in which nodes get similar representations during graph learning because of the information propagation in GNN. To tackle this, researchers propose different approaches:
DROPEDGE~\cite{rong2019dropedge} randomly prunes edges like a data augmentor that reduces the message passing speed, DEGNN~\cite{miao2021degnn,wang2021tree} applies connectivity aware decompositions that balance information propagation flow and overfitting issue, MADGap~\cite{chen2020measuring} measures the average distance ratio between intra-class and inter-class nodes which lower value ensures over-smoothing. However, these methods overlook networks' regional impact on oversmoothing. The \textbf{$k$-truss}~\cite{huang2014querying} algorithm primarily applies to community-based network operations to identify and extract various dense regions. It has been employed in different domains, such as high-performance computing~\cite{diab2020ktrussexplorer} and graph compression~\cite{akbas2017truss}. 

Our \TGS~model functions as a technique equipped with foundation graph pooling methods. It leverages the $k$-truss algorithm and edges' minimum node strength to provide networks' structural interconnectedness. Pruning highly dense connections helps to restrict excessive message passing paths to reduce oversmoothing in GNN models. We empirically justified it by experimenting with different graph topologies in section~\ref{sec: exp_and_res}. 

\section{Preliminaries}\label{sec:preliminaries}

This section discusses the fundamental concepts for GNN and pooling, and also formulate the oversmoothing problem including the essential components for our solution to this problem. We begin with discussing graph neural networks and graph pooling techniques. Then, define the issue, including the task. Finally, we delve into the foundation concept of our model ($k$-truss), which plays a crucial role in solving the problem. 

% This section discusses ..... We start with introducing GNN. Next, we describe the condition and definition of edge trussness and its utilization in graph processing algorithms.
\subsection{GNN and Graph Pooling}

\textbf{Graph Neural Network (GNN)}~\cite{scarselli2008graph} is an information processing framework that defines deep neural networks on graph data. Unlike traditional neural network architectures that excel in processing Euclidean data, GNNs are experts in handling non-Euclidean graph structure data. The principal purpose of GNN is to encode node, subgraph, and graph into low-dimension space that relies upon the graph's structure. In GNN, for each layer, $K$ in the range $1,2,...k$, the computational node aggregates~(\ref{eq:1}) messages $m_{N(v)}^{(k)}$ from its K-hop neighbors and updates~(\ref{eq:2}) its representation $h_v^{(k+1)}$ with the help of the $AGGREGATE$ function. %Finally, with $READOUT$ operation~(\ref{eq:3}) on nodes, the entire graph is embedded ($h_G$) for graph classification. 
\begin{flalign}
  m_{N(v)}^{(k)} = AGGREGATE^{(k)}(\{h_u^{(k)}, \forall u \in N(v)\}) \label{eq:1} \\
  h_v^{(k+1)} = UPDATE^{(k)}( h_v^{(k)}, m_{N(v)}^{(k)}) \label{eq:2}
  %h_G = READOUT( \{h_{v}^{(k)} | v \in V\} ) \label{eq:3}
\end{flalign}
In the context of graph classification, GNNs must focus on aggregating and summarizing information across the entire graph. Hence, the pooling methods come into play. 

\textbf{Graph Pooling.}~\cite{liu2022graph} 
Graph pooling performs a crucial operation in encoding the entire graph into a compressed representation. This process is vital for graph classification tasks as it facilitates capturing the complex network structure into a meaningful form in low-dimensional vector space. During the nodes' representation learning process at different layers, one or more pooling function(s) operate on them. These pooling layers are pivotal in enhancing the network's ability to generalize from graph data through effective graph summarization. In general, pooling operations are categorized into two types: Flat pooling and Hierarchical pooling.  

\textit{Flat pooling}~\cite{liu2022graph} is a straightforward graph readout operation. It simplifies the encoding by providing a uniform method to represent graphs of different sizes in a fixed size. 
\begin{flalign}
  h_G = READOUT( \{h_{v}^{(k)} | v \in V\} ) \label{eq:3}
\end{flalign}
 
\textit{Hierarchical pooling}~\cite{liu2022graph} 
 iteratively coarsens the graph and encodes comprehensive information in each iteration, reducing the nodes and edges of the graph and preserving the encoding. It enables the graph's representations to achieve short and long-sighted structural details. In contrast to Flat Pooling, it gives deeper insights into inherent graph patterns and relationships. 
 
 Between the two types of hierarchical graph pooling methods, the \textit{selection-based} methods emphasize prioritizing nodes by assigning them a score, aiming to retain the most significant nodes in the graph. They employ a particular $attention$ function for each node to compute the node importance. Based on the calculated scores, top $k$ nodes are selected to construct a pooled graph. The following equation gives a general overview of the top $k$ selection graph pooling method:
\begin{equation}
\begin{aligned}
S=score(G,X);~idx=topK(S,[\alpha \times N])\\
A^{(l+1)}=A_{idx,idx}
 \end{aligned}
 \label{equ: 4}
\end{equation}
where $S\in \mathbb{R}^{N\times 1}$ is the scores of nodes, $\alpha$ is the pooling ratio, and N is the number of nodes. Conversely, \textit{clustering-based} pooling methods form supernodes by grouping original graph nodes that summarize the original nodes' features. A cluster assignment matrix $S\in \mathbb{R}^{N\times K}$ using graph structure and/or node features are learned by the models. Then, nodes are merged into super nodes by $S\in \mathbb{R}^{N\times K}$ to construct the pooled graph at $(l+1)^{th}$ layer as follows
\begin{equation}\begin{aligned}\label{equ:5}
A^{(l+1)}=S^{(l)^T}A^{(l)}S^{(l)}\\
H^{(l+1)}= S^{(l)^T} H^{(l)}\end{aligned}
\end{equation} 
where $A\in{\mathbb{R^{N\times N}}}$ is the adjacency matrix and $H\in \mathbb{R}^{N\times d}$ is the feature matrix with $d$ dimensional node feature and $N$ is the number of nodes. Note that, the $AGGREGATE$, $UPDATE$ and $READOUT$  operations are different operational functions, commonly including $min$, $max$, $average$, and $concat$.

\subsection{Oversmoothing} \label{sec:oversmoothing}
According to~\cite{zhang2021node}, continual neighborhood aggregation of nodes' features gives an almost similar representation to nodes for an increasing number of layers $K$. simply, without considering the non-linear activation and transformation functions, the features converge as - 

\begin{equation}
    h^{\infty} = \hat{A}^{\infty}X, \quad \hat{A}_{i,j} = \frac{(d_i + 1)^{r}(d_j + 1)^{1-r}}{2m + n} \label{eq:oversmoothing1}
\end{equation}

\noindent where, $v_i$ and $v_j$ are source and target nodes, $d_i$ and $d_j$ are their degrees respectively, $\hat{A}$ is the final smoothed adjacency matrix and $r \in [0, 1]$ is the convolution coefficient. The equation~(\ref{eq:oversmoothing1}) shows for an infinite number of propagations, the final features are blended and only rely upon the degrees of target and source nodes. Furthermore, through spectral and empirical analysis ~\cite{chen2020simple} shows: \textit{nodes with higher-dree are more likely to suffer from oversmoothing}.

\begin{equation}
     h^{k}(j) = \sqrt{d_j + 1} (\sum_{i=1}^{n} \frac{\sqrt{d_j + 1}}{2m + n} x_i \pm \frac{\sum_{i=1}^{n}x_{i} (1-\frac{\lambda_{G}^2}{2})^k}{\sqrt{d_{j} +1}}) \label{eq:oversmoothing2}
\end{equation}
\noindent In the equation~(\ref{eq:oversmoothing2}), $\lambda_G$ is the spectral gap, $m$ is the number of edges, and $n$ is the number of nodes. It represents the features convergence relied upon the spectral gap~$\lambda_G$ and summation~$\sum_{i=1}^{n}$ of feature entries. When the number of layers $K$ goes to infinity, the second term disappears (after $\pm$). Hence, all vertices' features converge to steady-state for oversmoothing, which mainly depends on the nodes' degrees. 
\subsection{Problem Formulation}

% In graph analytics, classifying graphs is challenging due to their large size and complex structure. Graph sparsification-- a technique that reduces the number of graph connections by preserving crucial graph structures, is an emerging technique to address these challenges. 

This research aims to alleviate oversmoothing by effectively simplifying graphs to balance global and local connections, resulting in better graph classification results. Formally, A Graph is denoted as $G~=~(V, E, X)$, where ${V}$ is the set of nodes and ${E}$ is the set of edges. Symbol $X \in \mathbb{R}^{N \times d}$ represents the graph's feature matrix of dimension $d$, where $N = |V|$ is the number of nodes in $G$ and $x_{v} \in \mathbb{R}^d, x_{v} \in X~and~v \in V$ is a $d$ dimensional feature of a particular node in the graph. The neighborhood of a node $u$ is denoted as $N(u)$, and its degree is represented as $d(u) = |N(u)|$. For a dataset $D=(\mathbb{G}, Y)$ consisting of a set of graphs $\mathbb{G} = \{G_{1}, G_{2}\cdots G_{N}\}$, label pair $Y = \{Y_{1}, Y_{2}, ....Y_{N}\}$, our truss-based sparsification algorithm introduces a set of sparsified graphs as $\mathbb{G}_{S} = \{G_{S1}, G_{S2}....G_{SN}\}$. The algorithm is designed to remove redundant graph connections and retain the graph's essential structural information. Subsequently, This sparsified graphs set is analyzed using GNN models to learn a function $f:\mathbb{G}_{S} \rightarrow Y$ leveraging the reduced complexity of the graphs. The principal objective is to enhance the accuracy (Acc) of GNN models in graph classification tasks. 

% Reduced
% The simple objective function is - 
% \[\max_{G_S} \text{Acc}(f(\mathbb{G}_S), Y)\]

% predict the labels graph with different GNN models in an end-to-end manner. 

\subsection{$K$-truss}

Identifying and extracting cohesive subgraphs is a pivotal task in the study of complex networks. The $k$-truss subgraph extraction algorithm is instrumental as it isolates subgraphs based on a specific connectivity criterion. The root of the criterion is the term $support$, which refers to the enumeration of triangles in which an edge participates. The $support$ serves as the cornerstone to measure the cohesiveness of a subgraph. The following two definitions explain the criterion for extracting specific tightly interconnected subgraphs from a complex network.

\begin{theorem}\label{def:1}
\textit{\textbf{Support:} In graph $G = (V, E)$ , the support of an edge $e=(u, v) \in E$ is denoted as $sup_G(e)$ the number of triangles where e involves, i.e $~sup_G(e)~=~$ 
$ |\{~\Delta_{uvw}: w \in V~\}|$ }. 
\end{theorem}

\begin{theorem}\label{def:2}
\textit{\textbf{$k$-truss subgraph:} A subgraph $S = (V_S, E_s)$ where, $S \subseteq G,~V_s \subseteq V~and~E_S \subseteq E$ is a $k$-truss subgraph where every edge $e \in E_S$ has at least $k-2$ support, where $k \ge 2$}.
\end{theorem}

\noindent Notably, the concept of  $k$-truss is inherently dependent on the count of triangles within the graph, establishing that any graph can be considered a $2$-truss subgraph. The hierarchical structure of $k$-truss subgraphs implies that a $3$-truss subgraph is a subset of the $2$-truss subgraph (the original graph) denoted as $G_3 \subseteq G_2$. Similarly, $G_4 \subseteq G_3 \cdots G_k \subseteq G_{k-1}$.

\begin{theorem}\label{def:3}\textit{\textbf{Edge Trussness:}}
For a given graph $G$, for $k > 2$, an edge, $E(u, v)$, can exist in multiple $k$-truss subgraphs. The trussness of the edge, denoted as $T_{r}(u,v)$, is quantified from the highest $k$ value for which the edge is included in that subgraph. That is, $T_r(u,v)= k$ and $(u,v) \notin G_{k+1}$. 
\end{theorem}

\section{Truss based Graph Sparsification}\label{sec:methodology}
\begin{figure*}[t]
    \centering
    \includegraphics[width=\textwidth]{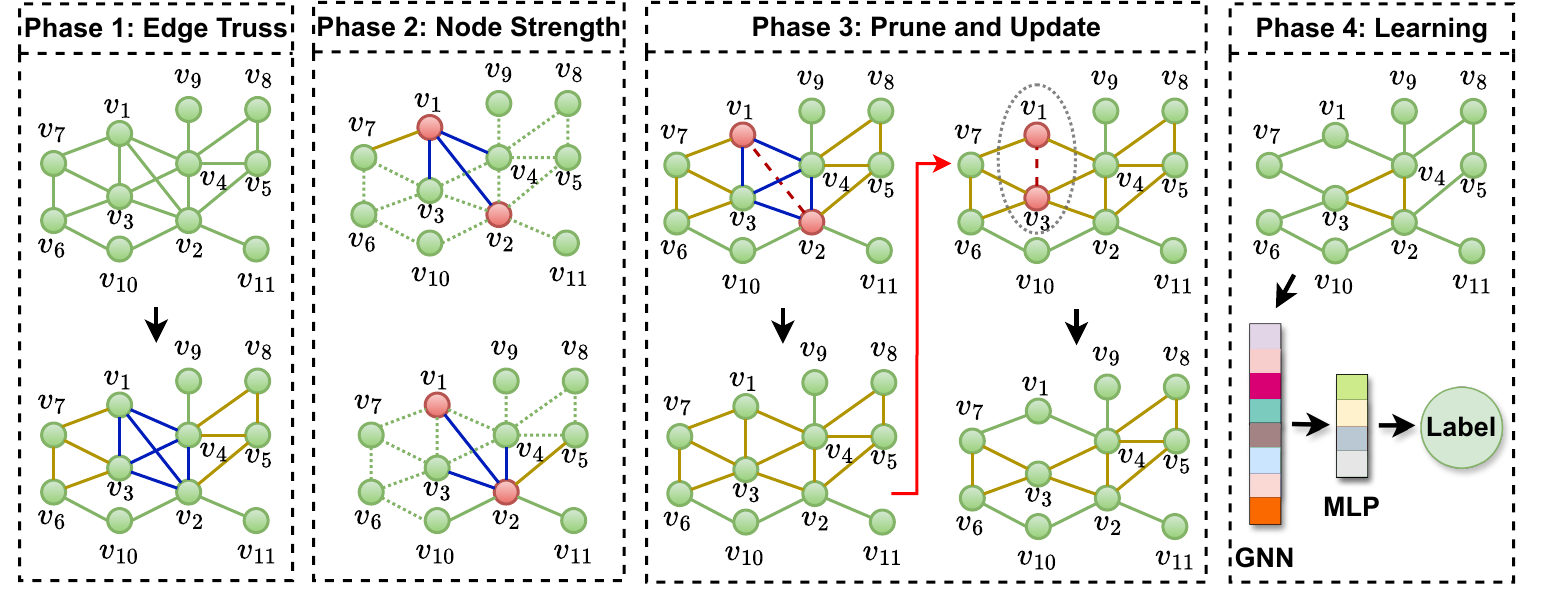}
    \caption{Architecture of the \TGS~$(\eta =3, \delta = 2.5)$. }
    \label{fig: diagram_figure}
\end{figure*}

In this section, we explain the proposed truss-based graph sparsification model (\TGS)~to overcome the oversmoothing problem for graph classification. In graph analytics, classifying graphs is challenging due to their large size and complex structure. Graph sparsification-- a technique that reduces the number of graph connections by preserving crucial graph structures, is an emerging technique to address these challenges. We aim to get an effective simplified graph that keeps essential short- and long-distance graph connections through graph sparsification, which produces the optimal graph classification result. The overall architecture of the proposed model is presented in Figure~\ref{fig: diagram_figure}. The model consists of 2 parts: truss-based graph sparsification and graph learning on the sparsified graph. We observe four phases to develop the truss-based graph sparsification framework.

\noindent\underline{Phase 1}: \textit{Compute edge trussness:} At first, we apply the k-truss decomposition algorithm on an unweighted graph to compute its edges' trussness as weight. 
% (line~\ref{ln:compute-edge-trussness}). 
Next, we split all edges into groups based on their truss values: high-truss edges and low-truss values for a given threshold $\eta$. 

% (line~\ref{ln:separate-high-truss-edges}). 

\noindent\underline{Phase $2$}: \textit{Measure node strenght:} \TGS~focuses on high-truss edges for sparsification. As high-truss edges have higher degrees, they massively contribute to the oversmoothing phenomenon~(Section~\ref{sec:oversmoothing}). Thus, strategic pruning of those edges helps to reduce oversmoothing. However, at the same time, important structural connections need to be maintained. To do so, we measure the minimum node strength of its two end nodes for each candidate high-truss edge, indicating the edge's surroundings' density status.

% ~(line~\ref{ln:compute-min-node-strength}). 

\noindent\underline{Phase 3}: \textit{Prune and update:} When that minimum value exceeds the standard density assuring threshold, we prune the edge from the graph and update all the edge's trussness values. Due to the cascading effect of the network, pruning affects other edges' trussness. Therefore, edge trussness needs to be updated after each pruning operation.
% (lines~\ref{ln:pruning-the-edge}-~\ref{ln:update-edge-trussness}). 
The process continues the pruning step until all high-truss edges are examined.

% ~(line~\ref{ln:start-sparsification}-~\ref{ln:end-of-sparsification}). 

\noindent\underline{Phase 4}: \textit{Learning:} At the end of the sparsification, we first feed the processed graph to GNN models for graph learning. Finally, we experiment the entire graph's representation with a multi-layer perceptron (MLP) network.

Note that our model follows some strategies during the graph sparsification steps: $(a)$ sorts the high-truss edges in descending order to prune more dense regions' edges earlier, and $(b)$ examines each edge only once. Removing an edge from the graph might affect other edges; then, in further exploration, one edge might satisfy the pruning condition. The phenomenon negligibly happens as \TGS~starts to prune from more dense edges. Hence, the technique avoids recursion.

\subsection{Dense Region Identification}
To learn the structure of the graph, GNN applies message passing between nodes through edges. Through repeated message passing, nodes in the dense regions get similar neighbors' feature information, which causes oversmoothing. As a result, the features of those regions' nodes become indistinguishable. Many different density measures exist, including $k$-truss, $k$-core, and $k$-edge. This paper uses $k$-truss, defined based on triangle connectivity, to identify the dense regions. 
% As opposed to primitive vertices/edges, the higher-order
% graph motif, triangle, is exploited as building blocks to quantify the strong and stable relationships within the graph.

\begin{algorithm}[h!]
\SetAlgoLined
  \KwInput{A Graph $G = (V, E)$}
  \KwOutput{A graph $G_T$, where edge trussness $T_r (e)$ as weight for each $e \in E$}
\DontPrintSemicolon
% \SetAlgoNoEnd
Compute$~sup_{G}(e),\; \forall e \in E$ \\
$Sort(e \in E,~item = sup_{G}(e))$~// in non-decreasing order \\ 
$k \gets 2~$ ; $G_T \gets G.copy()$ \\
\While{$\exists e \in E, sup_G(e) \le (k-2)$} 
{
  \label{wlp: 1}
  $e^{*}(u,v) \gets argmin_{e \in E}~sup_{G}(e)$ // assume w.o.l.g,~$d(u) \le d(v)$ \label{ln:5}\\
  
  \ForEach{ $w \in N(u) \cap N(v)$ \textbf{and} $~e^{*} =(u,v) \in E$}
  {
    $sup_G(u, w) \gets sup_G(u, w) - 1$ \\
    $sup_G(v, w) \gets sup_G(v, w) - 1$ \\

    Reorder$(u,w)$ and $(v,w)$\\ \label{ln:10}
  }
  $G_T[u][v][W] \gets T_r(e^{*}) \gets k$;\\
  remove~$e^{*}$~from $E$\\ \label{ln:12}
}
\If{$\exists e \in E$}
{
\label{ln:14}
$k \gets k + 1$\\
\textbf{goto} the while-loop~(line~\ref{wlp: 1})\\
}
\label{ln:17}

\Return $G_T$
 \caption{Computing Edge Trussness}
 \label{alg:alg1}
\end{algorithm}

Our approach employs a truss-decomposition~\cite{akbas2017truss} algorithm,
as detailed in Algorithm~\ref{alg:alg1}, to compute edge trussness and
discover all k-trusses from G.
At first, it takes an unweighted graph as input and computes the supports of all edges. Then initialize the value of $k$ as 2 and select the edge $e^{*}$ with the lowest support (line~\ref{ln:5}). Next, the value $k$ is assigned as edge weight $W$, and the edge is removed (line~\ref{ln:12}). Removing an edge decreases other edges supports. Hence, we reorder edges according to their new support values (line~\ref{ln:10}). The process continues until the edges that have support no greater than $(k-2)$ are removed from the graph. Next, the algorithm checks whether any edge exists to access or not. If one or more exist(s), it increments $k$ by one and goes to the line~\ref{wlp: 1} again to measure their trusseness (line ~\ref{ln:14}-\ref{ln:17}). Edge trussness facilitates understanding the highest dense region within which the edge exists. After calculating the edge trussness, to identify highly dense areas, \TGS~separates the edges in $G_T$ into two sets: High-Truss Edges and Low-Truss Edges. Following condition~(\ref{cond:1}), it compares all edges' trussness with the given threshold value, $\eta$, and determines the High Truss Edges $E_H$. For example, in Figure~\ref{fig: diagram_figure}, given $\eta = 3$, the blue ($T_r(E) = 4$) and golden ($T_r(E) = 3$) colored edges are high-truss edges. Pruning $\{E \in E_H\}$ reduces the load of high-degree nodes in dense regions, which assists in mitigating oversmoothing in GNN.

\begin{condition}\label{cond:1}
    \textit{\textbf{High Truss Edges $E_H$:} In any graph for a specific variable $\eta$, if an edge's trussness value is greater than or equal to $\eta$ then the edge is considered as a high truss edge and their set is denoted as $E_H$. }
\end{condition}

\subsection{Pruning Redundant Edges}
Ascertaining the high-truss edges is crucial for understanding the density level in different parts of the graph. However, directly pruning these edges may break up essential connectivity between nodes. For example, low-degree nodes could be connected with a dense region node, and pruning an incident high-truss edge may not provide adequate information to that low-degree node. To balance the connectivity between nodes, we determine the nodes' strength of edge high-truss edges and then proceed to the next step. 
To measure nodes' ($n \in E,~E \in E_H$) strength, \TGS~calculates the average trussness $\bar{T_{N(n)}}, n \in V$. This score ensures the density depth of a node and implies its important connectivity.

\begin{theorem}\label{def:4}
The strength of a node is measured as the summation of all of its incident edge weights. However, In this research, \textbf{node strength} is applied as the average of nodes' incident edges' trussness. 
\begin{equation}\begin{aligned}\label{eq:6}
    \bar{T_{N(n)}} \gets \frac{1}{|N(n)|} \sum_{u\in N(n)} T_r(n,u)  \\
    % \bar{T_{N(v)}} \gets \frac{1}{|N(v)|} \sum_{n\in N(v)} T_r(v,n) 
\end{aligned}
\end{equation} 
\end{theorem}

 For a candidate edge $E=(u,v)$, after measuring the node strength of $u$~and~$v$~(\ref{eq:6}) , their minimum value~(\ref{eq:7}) has been taken. Notably, a node may be included in different $k$-truss subgraphs. Hence, its neighborhood's trussness provides more connectivity information.  The minimum node strength of an edge's two endpoints signifies the least density of its surroundings. As we aim to reduce the density of highly connected regions to combat oversmoothing, \TGS~follows a technique to decide to prune edges. For this purpose, the minimum node strength of the edge $E$ is compared to a threshold $\delta$. In condition~(\ref{cond:2}), This comparison ensures the edge's presence in a prunable dense region. The condition indicates that if any end of the candidate edge is sparse $\bar{T_{N(E)}} < \delta$, \TGS~avoids cutting it because that connection serves as an essential message-passing medium in the GNNs aggregation step. In contrast, when the minimum score equals or exceeds the value of $\delta$, we assume the edge is part of a highly dense region, and there is a high chance of excessive messages passing between that region's nodes. That may cause them to blend their representations, leading to oversmoothing during graph learning through GNN (section~\ref{sec:oversmoothing}). From the condition, the model understands which edges contribute to undesirable density levels that foster oversmoothing in GNNs.  

\begin{equation}\begin{aligned}
        \bar{T_{N(E)}}  \gets minimum(\bar{T_{N(u)}}, \bar{T_{N(v)}}) \label{eq:7} 
\end{aligned}\end{equation}

\begin{algorithm}[t]
\SetAlgoLined
  \KwInput{ A Graph $(G)$, threshold $\delta$, cutoff $\eta$ }
  \KwOutput{A Sparsified Graph  $G_S \subset G$}
\DontPrintSemicolon
% \SetAlgoNoEnd
{$G_T = Computing~Edge~Trussness(G)$} \label{ln:compute-edge-trussness} \\
{$E_H = \{ \forall (u,v) \in E, T_r(u,v) \ge \eta \}$ } \label{ln:separate-high-truss-edges}\\
{$E_H \gets Sort(E_H,reverse = True)$ } \label{ln:sort-high-truss-edges} \\
{$n \gets length(E_H)$} \\
{$G_S \gets G_T$} \\
\For{$r\gets1$ \KwTo $n$ }
{
    \label{ln:start-sparsification}
    \ForEach{ $E(u,v) \in E_H$}
    {   
        Compute $\bar{T_{N(E)}}$ from Equations~(\ref{eq:6}) and~(\ref{eq:7}) \label{ln:compute-min-node-strength} \\
        \If{$\bar{T_{N(E)}} \ge \delta$}
            {   \label{ln:checking-for-pruning}
                $G_S \gets G_T \backslash E(u,v) $ // Edge Cut \\
                $E_H \gets E_H \backslash E(u,v)$ \label{ln:pruning-the-edge} // Reduce High Truss Edge List \\
                $G_S \gets UpdateTr(G_S)$ \label{ln:update-edge-trussness} \\
            }  
    }
}
\label{ln:end-of-sparsification}

\Return $G_S$
 \caption{Truss-based Graph Sparsification (\TGS)}
 \label{alg:alg2}
\end{algorithm}

\begin{condition}\label{cond:2}
An Edge $e =(u,v)$ is eligible for pruning when the minimum average neighborhood edge weight between $u$ and $v$ equals or exceeds the threshold $\delta$. 
\begin{align}
        \bar{T_{N(E)}} \ge \delta \label{eq: 9}
\end{align}
\end{condition}

\noindent For example, at the lower-left in \underline{phase $3$}~(in Figure~\ref{fig: diagram_figure}),  the edge ($v_{2}$,  $v_{10}$), where the degrees of $v_{2}$ and $v_{10}$ are $5$ and $2$, respectively. The node strengths, $\bar{T_{N(v_{2})}}$ is $\{(3\times3)+(2\times2)\}/5 = 2.6$, and $\bar{T_{N(v_{10})}}$ is $\{(2+2)\}/2 = 2$. Given $\delta = 2.5$, and the minimum node strength, $\bar{T_{N(v_2, v_{10})}} = \bar{T_{N(E)}} = min~( 2.6,~ 2)~=~ 2$. Hence, the pruning condition is unsatisfactory, and the edge will stand in the graph. If \TGS~pruned the edge, the neighborhood of $v_{10}$ would be sparser than before and miss its crucial global information. On the other hand, at the upper-right in \underline{phase $3$}, in context of $E=(v_1$, $v_3)$, $\bar{T_{N(v_1)}} = 3$ and $\bar{T_{N(v_3)}} = 3$. Hence, comparing to the value of $\delta$ they are already in the dense region and $\bar{T_{N(v_1, v_3)}} = 3 \ge 2.5$. In this case, the pruning will help to prevent blended node representation in GNN, especially between highly interconnected subgraphs. According to our model, it considers the nodes will still stay in enough dense regions to receive meaningful local and global neighborhood information after pruning.

The Algorithm~\ref{alg:alg2} represents the \TGS~model.  Lines ~(\ref{ln:compute-edge-trussness}-~\ref{ln:sort-high-truss-edges}) identify the dense regions and ensure high-truss edges of the network while lines~(\ref{ln:start-sparsification}-\ref{ln:end-of-sparsification}) demonstrate the pruning of noisy high-truss edges in the network. Algorithm $UpdateTr$ in line~\ref{ln:update-edge-trussness}~(similar to section~$4.2$, \cite{huang2014querying}), updates all edges' trussness after each pruning step.

 \subsection{Algorithm Complexity} The complexity of measuring edge trussness is $O(E$ $\sqrt(E))$ and the updateTr algorithms complexity is O(E). As we explore all high truss edges in the algorithm, the exploration complexity is O(E) in the worst case. Hence, the Algorithms complexity is  $O(E(O(E)) + O(E\sqrt(E))$ $= O(E^2) + O(E\sqrt(E))$ = $O(E^2) $ in the worst case. Although it seems highly complex the real-world datasets are not hardly dense. In addition, due to updating the edges trussness score many high-truss edges are removed before examination.

\begin{table}[t]
\centering
\caption{Datasets' Statistics}
\label{table: tbl_data}
\begin{tabular}{|c|c|c|c|c|}
\hline
\textbf{\cellcolor{gray!15} Datasets} & \cellcolor{gray!15} \textbf{\# Graphs} & \cellcolor{gray!15} \textbf{Avg \# $|V|$} & \cellcolor{gray!15} \textbf{Avg \# $|E|$} & \cellcolor{gray!15} \textbf{\# Classes} \\ \hline
\textbf{\cellcolor{gray!15} PROTEINS} & 1113 & 39.06 & 72.82 & 2 \\ \hline
\textbf{\cellcolor{gray!15} NCI1} & 4110 & 29.87 & 32.30 & 2 \\ \hline
\textbf{\cellcolor{gray!15} NCI109} & 4127 & 29.68 & 32.13 & 2 \\ \hline
\textbf{\cellcolor{gray!15} PTC} & 344 & 25.56 & 25.96 & 2 \\ \hline
\textbf{\cellcolor{gray!15} DD} & 1178 & 284.32 & 715.66 & 2 \\ \hline \hline
\textbf{\cellcolor{gray!15} IMDB-B} & 1000 & 19.77 & 96.53 & 2 \\ \hline
\textbf{\cellcolor{gray!15} IMDB-M} & 1500 & 13.00 & 65.94 & 3 \\ \hline
\textbf{\cellcolor{gray!15} REDDIT-B} & 2000 & 429.63 & 497.75 & 2 \\ \hline
\end{tabular}
\end{table}

\begin{table*}[ht!]

\caption{Result Table (in \%). Results that achieve at least 0.5\% gain over the counterpart model we mark in bold. We show a model-by-model comparison. 
% The models before and after the row divider are state-off-the-art graph pooling models and baseline GNN models for graph classification. Here, M and B after the dataset name indicate 'MULTI' and 'BINARY'
}
\label{table:tbl_result}
% \usepackage{multirow}
\centering
\begin{tabular}{|c|c|ccccc|ccc|}
\hline
 \multicolumn{1}{|}{}  &  \multicolumn{1}{l|}{} & \multicolumn{5}{c|}{\cellcolor{gray!15} \textbf{ Biomedical Dataset}} & \multicolumn{3}{c|}{\cellcolor{gray!15}  \textbf{ Social Network Dataset}} \\ \cline{3-10}
\multicolumn{2}{|c|}{ \textbf{Backbones}} & \multicolumn{1}{|c|}{\cellcolor{gray!15}\textbf{ PROTEINS}} & \multicolumn{1}{c|}{\cellcolor{gray!15}~~\textbf{ NCI1}~~} & \multicolumn{1}{c|}{\cellcolor{gray!15}~\textbf{ NCI109}~} & \multicolumn{1}{c|}{\cellcolor{gray!15}~~\textbf{ PTC}~~} & \cellcolor{gray!15} ~~~\textbf{DD}~~ & \multicolumn{1}{c|}{\cellcolor{gray!15} \textbf{ IMDB-B}} & \multicolumn{1}{c|}{\cellcolor{gray!15} \textbf{ IMDB-M}} & \cellcolor{gray!15} \textbf{REDDIT-B} \\ \hline
\multirow{2}{*}{SAGPool} & \multicolumn{1}{l|}{ \cellcolor{gray!15} Original} & \multicolumn{1}{c|}{72.68} & \multicolumn{1}{c|}{70.10} & \multicolumn{1}{c|}{67.37} & \multicolumn{1}{c|}{57.14} & 77.31 & \multicolumn{1}{c|}{71.60} & \multicolumn{1}{c|}{44.87} & 77.55 \\ 
 & \cellcolor{gray!15} \TGS & \multicolumn{1}{c|}{\textbf{73.84}} & \multicolumn{1}{c|}{\textbf{70.72}} & \multicolumn{1}{c|}{\textbf{67.78}} & \multicolumn{1}{c|}{\textbf{58.00}} & \textbf{80.67} & \multicolumn{1}{c|}{\textbf{75.30}} & \multicolumn{1}{c|}{44.67} & \textbf{79.05} \\ \hline
\multirow{2}{*}{GMT} & \multicolumn{1}{l|}{\cellcolor{gray!15} Original} & \multicolumn{1}{c|}{70.63} & \multicolumn{1}{c|}{60.29} & \multicolumn{1}{c|}{48.86} & \multicolumn{1}{c|}{50.86} & 65.88 & \multicolumn{1}{c|}{71.10} & \multicolumn{1}{c|}{47.40} & 70.95 \\ 
 & \cellcolor{gray!15} \TGS & \multicolumn{1}{c|}{\textbf{71.25}} & \multicolumn{1}{c|}{\textbf{61.56}} & \multicolumn{1}{c|}{\textbf{52.32}} & \multicolumn{1}{c|}{51.14} & \textbf{69.50} & \multicolumn{1}{c|}{\textbf{74.50}} & \multicolumn{1}{c|}{\textbf{47.93}} & 71.25 \\ \hline
\multirow{2}{*}{DiffPool} & \multicolumn{1}{l|}{\cellcolor{gray!15} Original} & \multicolumn{1}{c|}{\textbf{71.10}} & \multicolumn{1}{c|}{66.96} & \multicolumn{1}{c|}{67.60} & \multicolumn{1}{c|}{46.29} & 75.31 & \multicolumn{1}{c|}{63.90} & \multicolumn{1}{c|}{\textbf{44.80}} & 80.92 \\  
 & \cellcolor{gray!15} \TGS & \multicolumn{1}{c|}{69.82} & \multicolumn{1}{c|}{\textbf{68.30}} & \multicolumn{1}{c|}{67.20} & \multicolumn{1}{c|}{\textbf{55.14}} & \textbf{78.75} & \multicolumn{1}{c|}{\textbf{64.80}} & \multicolumn{1}{c|}{42.67} & \textbf{82.46} \\ \hline
\multirow{2}{*}{DMon} & \multicolumn{1}{l|}{\cellcolor{gray!15} Original} & \multicolumn{1}{c|}{75.45} & \multicolumn{1}{c|}{74.20} & \multicolumn{1}{c|}{72.61} & \multicolumn{1}{c|}{53.71} & 79.32 & \multicolumn{1}{c|}{74.00} & \multicolumn{1}{c|}{49.53} & 84.95 \\  
 & \cellcolor{gray!15} \TGS & \multicolumn{1}{c|}{75.80} & \multicolumn{1}{c|}{\textbf{74.74}} & \multicolumn{1}{c|}{\textbf{73.77}} & \multicolumn{1}{c|}{\textbf{56.00}} & \textbf{80.42} & \multicolumn{1}{c|}{\textbf{74.90}} & \multicolumn{1}{c|}{49.60} & \textbf{85.85} \\ \hline
\multirow{2}{*}{MinCut} & \cellcolor{gray!15} Original & \multicolumn{1}{c|}{73.75} & \multicolumn{1}{c|}{72.77} & \multicolumn{1}{c|}{73.28} & \multicolumn{1}{c|}{56.28} & 76.63 & \multicolumn{1}{c|}{71.40} & \multicolumn{1}{c|}{51.06} & 76.85 \\ 
 & \cellcolor{gray!15} \TGS & \multicolumn{1}{c|}{73.84} & \multicolumn{1}{c|}{\textbf{73.58}} & \multicolumn{1}{c|}{73.04} & \multicolumn{1}{c|}{\textbf{58.57}} & \textbf{77.31} & \multicolumn{1}{c|}{\textbf{72.80}} & \multicolumn{1}{c|}{51.20} & 77.05 \\ \hline
\multirow{2}{*}{AdamGNN} & \cellcolor{gray!15} Original & \multicolumn{1}{c|}{78.12} & \multicolumn{1}{c|}{47.36} & \multicolumn{1}{c|}{65.16} & \multicolumn{1}{c|}{60.00} & 70.63 & \multicolumn{1}{c|}{72.48} & \multicolumn{1}{c|}{49.53} & OOM \\ 
 & \cellcolor{gray!15} \TGS & \multicolumn{1}{c|}{\textbf{81.77}} & \multicolumn{1}{c|}{47.36} & \multicolumn{1}{c|}{\textbf{67.81}} & \multicolumn{1}{c|}{\textbf{62.86}} & \textbf{74.25} & \multicolumn{1}{c|}{\textbf{77.60}} & \multicolumn{1}{c|}{\textbf{50.57}} & OOM \\ \hline
\multirow{2}{*}{HGP-SL} & \cellcolor{gray!15} Original & \multicolumn{1}{c|}{74.64} & \multicolumn{1}{c|}{73.33} & \multicolumn{1}{c|}{72.87} & \multicolumn{1}{c|}{56.00} & 72.35 & \multicolumn{1}{c|}{72.90} & \multicolumn{1}{c|}{49.47} & OOM \\  
 & \cellcolor{gray!15} \TGS & \multicolumn{1}{c|}{74.28} & \multicolumn{1}{c|}{73.38} & \multicolumn{1}{c|}{\textbf{74.22}} & \multicolumn{1}{c|}{\textbf{57.43}} & \textbf{73.19} & \multicolumn{1}{c|}{\textbf{76.10}} & \multicolumn{1}{c|}{\textbf{49.80}} & OOM \\ \hline \hline
\multirow{2}{*}{GIN-0} & \cellcolor{gray!15} Original & \multicolumn{1}{c|}{73.42} & \multicolumn{1}{c|}{81.70} & \multicolumn{1}{c|}{74.99} & \multicolumn{1}{c|}{68.86} & 74.58 & \multicolumn{1}{c|}{73.00} & \multicolumn{1}{c|}{47.60} & 73.60 \\ 
 & \cellcolor{gray!15} \TGS & \multicolumn{1}{c|}{73.84} & \multicolumn{1}{c|}{\textbf{82.50}} & \multicolumn{1}{c|}{75.08} & \multicolumn{1}{c|}{\textbf{69.43}} & \textbf{75.93} & \multicolumn{1}{c|}{\textbf{78.10}} & \multicolumn{1}{c|}{\textbf{52.53}} & \textbf{74.35} \\ \hline
\multirow{2}{*}{GIN-e} & \cellcolor{gray!15} Original & \multicolumn{1}{c|}{73.12} & \multicolumn{1}{c|}{81.85} & \multicolumn{1}{c|}{75.93} & \multicolumn{1}{c|}{68.28} & \textbf{76.44} & \multicolumn{1}{c|}{73.80} & \multicolumn{1}{c|}{49.33} & 73.55 \\ 
 & \cellcolor{gray!15} \TGS & \multicolumn{1}{c|}{73.39} & \multicolumn{1}{c|}{\textbf{82.50}} & \multicolumn{1}{c|}{\textbf{76.97}} & \multicolumn{1}{c|}{68.29} & 74.75 & \multicolumn{1}{c|}{73.90} & \multicolumn{1}{c|}{\textbf{53.40}} & \textbf{74.55} \\ \hline 
\multirow{2}{*}{GCN} & \cellcolor{gray!15} Original & \multicolumn{1}{c|}{68.29} & \multicolumn{1}{c|}{71.41} & \multicolumn{1}{c|}{69.61} & \multicolumn{1}{c|}{52.00} & 64.87 & \multicolumn{1}{c|}{75.20} & \multicolumn{1}{c|}{50.00} & 82.30 \\ 
 & \cellcolor{gray!15} \TGS & \multicolumn{1}{c|}{\textbf{70.45}} & \multicolumn{1}{c|}{\textbf{72.24}} & \multicolumn{1}{c|}{69.69} & \multicolumn{1}{c|}{\textbf{52.57}} & \textbf{73.10} & \multicolumn{1}{c|}{75.50} & \multicolumn{1}{c|}{\textbf{50.60}} & \textbf{84.30} \\ \hline
\end{tabular}
\end{table*}

\section{Experiment Design and Analysis}\label{sec: exp_and_res} 

This section validates our technique on different real-world datasets by applying standard graph pooling models. First, we provide an overview of the datasets. Then, we briefly describe the parameters of various methods. Finally, we compare the performance of our \TGS~algorithm's enhancement with the original baselines in the graph classification tasks including analysis of parameters, deeper networks, and ablation study.

\subsection{Datasets and Baselines}
We experiment with our model on eight different TU Dortmund~\cite{morris2020tudataset} datasets: Five of them are biomedical domain:~\textbf{PROT}EINS,\textbf{ NCI1, NCI109, PTC, and DD} and three of them from social network domain:~\textbf{IMDB-B}INARY, \textbf{IMDB-M}ULTI, and \textbf{REDDIT-B}INARY. We extend the \TGS~algorithm with seven state-of-the-art backbone graph pooling models. Among them, three are node clustering-based pooling methods: \textbf{Diff}Pool~\cite{ying2018hierarchical}, \textbf{DMon}Pool~\cite{tsitsulin2023graph} and \textbf{M}i\textbf{nCut}Pool~\cite{bianchi2020spectral}. Two models, \textbf{SAG}Pool~\cite{lee2019self} and \textbf{HGP-SL~\cite{zhang2019hierarchical}}, utilize a node selection approach for pooling the graphs. Of the remaining two, one learns graph representation through flat-pooling (\textbf{GMT~\cite{baek2021accurate}}), and another one utilizes an adaptive pooling approach by applying both node selection and clustering for the pooling procedure:~\textbf{Ad}am\textbf{G}NN~\cite{zhong2022multi}. 
% Most of these datasets consist of relatively sizable network information. Hence, it is compatible to evaluate them with deep graph learning models. 
We report the statistics of the datasets in the Table~\ref{table: tbl_data}.

\subsection{Experimental Settings} \label{subsec:experiment_setting}
To compare fairly, we executed the existing standard implementations of the baselines and incorporated them with our model. For evaluation, we split the datasets into $80\%$ for training, $10\%$ for validation, and $10\%$ testing. Mostly, we stopped learning early for $50$ consecutive same validation results in training. We measured the performance using the accuracy metric by ruining each model 10 times for 10 random seeds and reported their mean. The batch size was 128 for most of the models. The effectiveness of our pruning method mostly depends on two crucial parameters: the cutoff parameter $\eta$ and the edge pruning threshold $\delta$. For all experiments, we set $\eta = 3$, which means any edge with a trussness score below $3$ cannot be pruned from the graph. On the other hand, we experimented with various $\delta$ values across the datasets. Specifically, we used $\delta$ values of $\{3, 4, 5, 6, 7\}$ for IMDB-BINARY, IMDB-MULTI, and $\{3, 3.5, 4\}$ for REDDIT-BINARY datasets. For PROTEINS and DD datasets $\delta$ was set as $\{3, 3.25, 3.5,$ $3.75,~and~4\}$  and for NCI1, NCI109, we used $\delta$ values of \{ 2.5, and 3\} while for PTC only $2.5$.

\subsection{Result Analysis}
Table~\ref{table:tbl_result} reports the experiment results, providing a comparative analysis between our established model and original baselines across various datasets. \TGS~integrated with backbone graph pooling models consistently outperforms the baselines, and demonstrates its robustness in graph classification tasks. On selection-based models, with the incorporation of the SAGPool model, \TGS~achieves a 1.5-5.5\% gain$(\mathbb{G})$~(\ref{eq:gain}) over the original models. Notably, on DD and IMDB-BINARY datasets, the gains are 3.34\% and 5.17\%, respectively. In the experiment with the HGP-SL model, \TGS~attains a sustainable improvement of nearly 4.5\%  on the IMDB-BINARY dataset and on the PTC dataset, which is over 2.5\%. Adapting \TGS~along with the flat pooling model GMT acquires a significant gain (nearly 7\%) over the NCI109 dataset and maintains consistent performance on other datasets. 

In experiments with cluster-based modes, \TGS~equipped to Diffpool model achieves a magnificent accuracy gain on the PTC dataset, which is nearly 19\%. It also demonstrates strong performance with the DMonPool model over all datasets. Notably, it achieves the highest accuracy on the REDDIT-BINARY dataset, which is 85.75\% %. \TGS~shows Equally impressive performance with another clustering-based pooling model, MinCutPool, which acquires the best accuracy for the NCI1 dataset, which is 73.58\%. In the case of the AdamGNN, \TGS~gains significant amounts from most of the datasets. Specifically, on IMDB-BINARY, it gets near 7\% accuracy gain, whereas on the PROTEINS dataset, the gain is more than 4.5\%. in Table~\ref{table:tbl_result}, we omit gain for brevity, and OOM means the experiment experienced \textbf{O}ut \textbf{O}f \textbf{M}emory.

\begin{equation}
\label{eq:gain}
Gain(\mathbb{G}) =  \frac{\text{\TGS} -\text{Original} }{\text{Original}}  \times 100\% 
\end{equation}

\textbf{Extended Experiment}: In addition to the pooling method, we incorporate the \TGS~model with the fundamental GNN models: two versions of graph isomorphic networks (GIN-0 and GIN-$\epsilon$) and the simple graph convolution network for graph classification. During the experiment with GIN networks, we follow 10-fold cross-validation to evaluate the validity of our model. On the other hand, we assess the GCN as other pooling models (section~\ref{subsec:experiment_setting}). In most contexts~(in Table~\ref{table:tbl_result}), our technique outperforms these models on every dataset. Especially on the PTC and IMDB-BINARY datasets,~\TGS(GIN-0) achieves the highest accuracy scores of $69.43\%$ and $78.10\%$, respectively. Additionally, \TGS~with the backbone GCN model, attains the overall second-highest accuracy on the REDDIT-BINARY data, with $84.30\%$. 

% \noindent\textbf{Extended Experiment}: In addition to the pooling method, we compare our integrated model to the fundamental GNN models: two versions of graph isomorphic networks (GIN-0 and GIN-$\epsilon$) and the simple graph convolution network (GCN) for graph classification. In experiments with GIN networks, we follow 10-fold cross-validation to evaluate the validity of our model. On the other hand, we assess the GCN as other pooling models (as~\ref{subsec:experiment_setting}). In most contexts~(in Table~\ref{table:tbl_result}), our technique outperforms these models on every dataset. 
\begin{figure*}[t]
    \centering
    \begin{subfigure}[h]{0.24\textwidth}
        \centering
        \includegraphics[width=\textwidth]{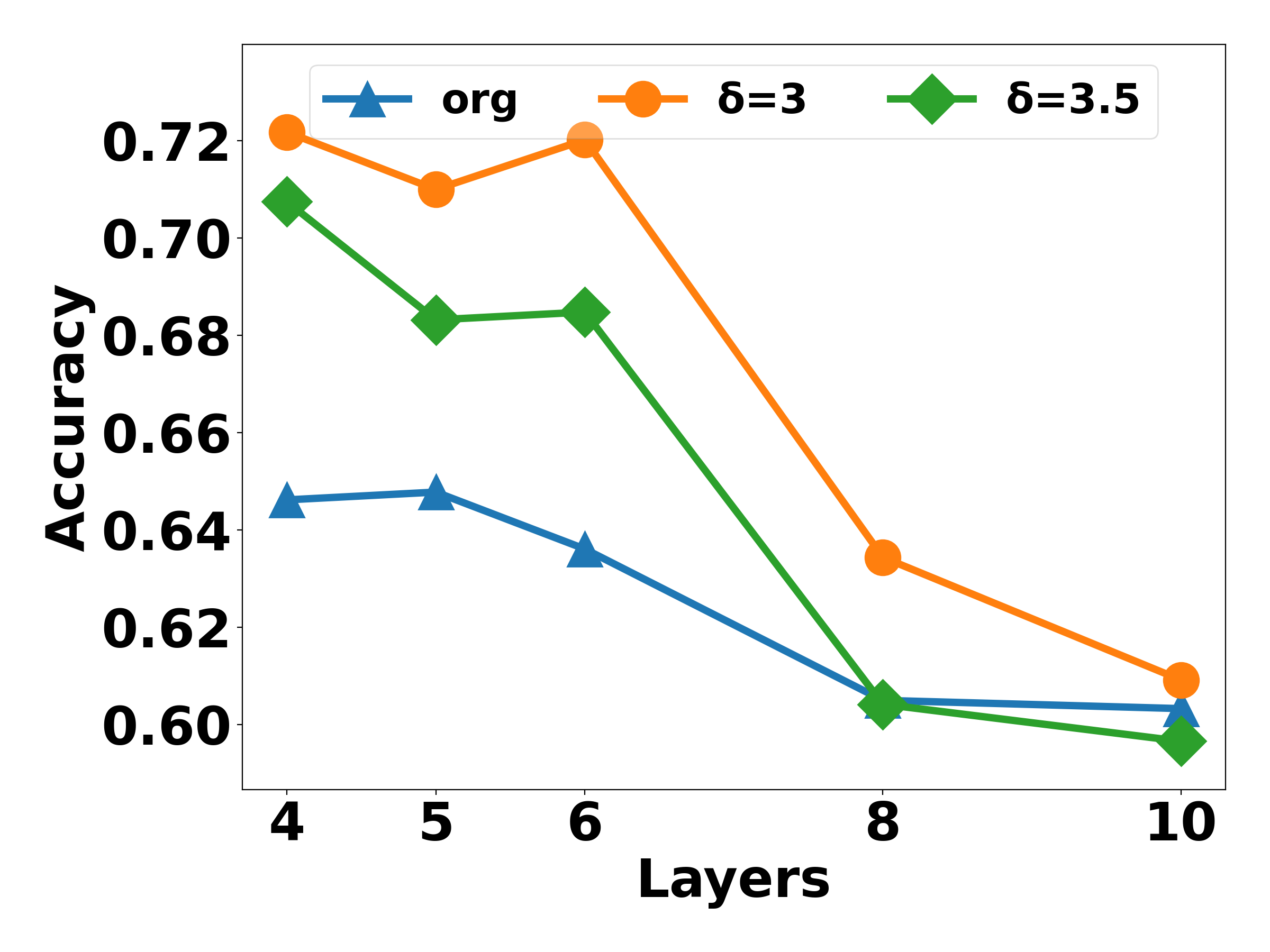}
        \caption{DD(GCN)}
        \label{fig:dd_gcn}
    \end{subfigure}%
     ~
    \begin{subfigure}[h]{0.24\textwidth}
        \centering
        \includegraphics[width=\textwidth]{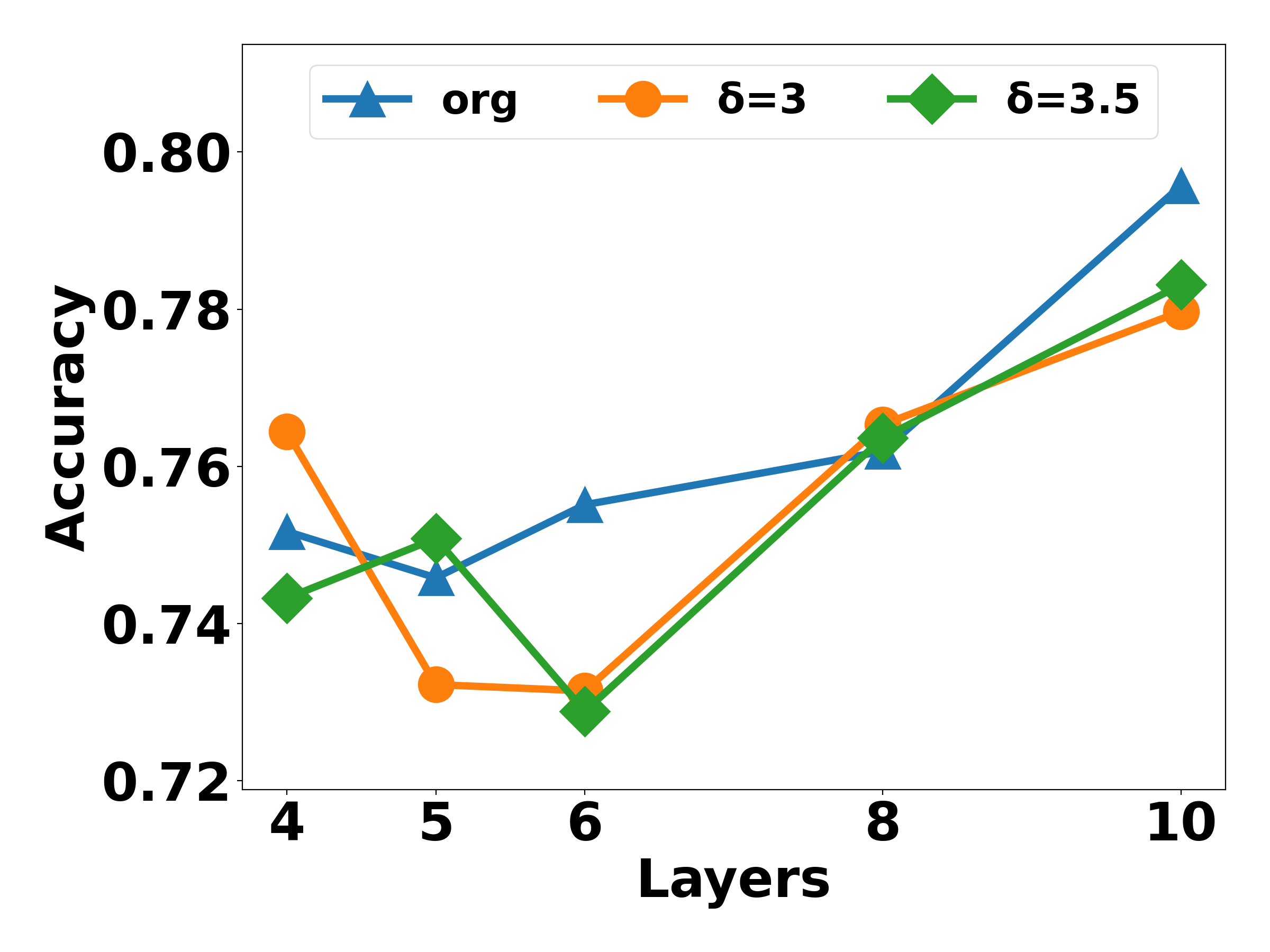}
        \caption{DD(GIN-$0$)}
        \label{fig:dd_gin_0}
    \end{subfigure}%
   ~
    \begin{subfigure}[h]{0.24\textwidth}
        \centering
        \includegraphics[width=\textwidth]{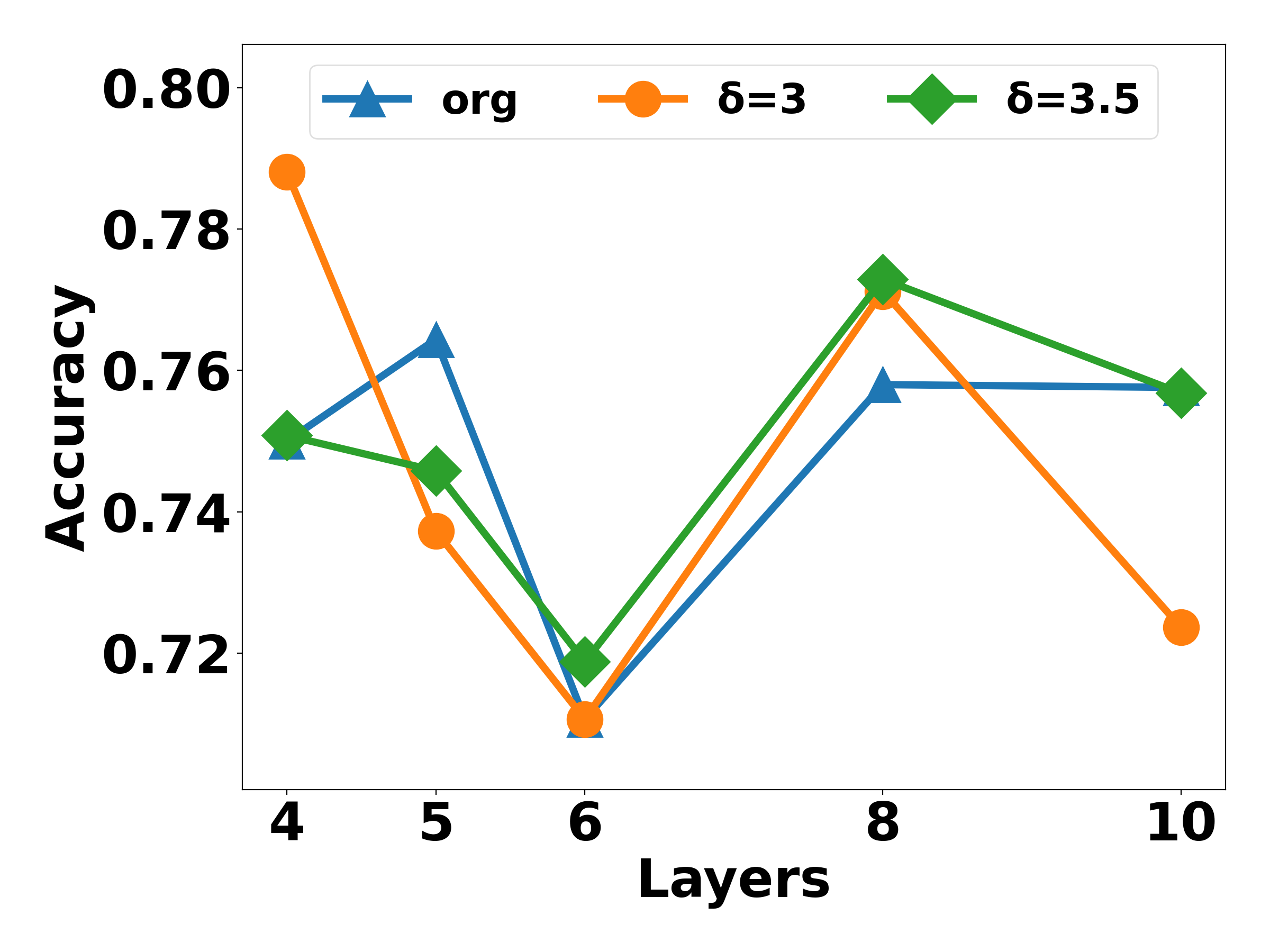}
        \caption{DD(GIN-$\epsilon$)}
        \label{fig:dd_gin_e}
    \end{subfigure}%
    ~
    \begin{subfigure}[h]{0.24\textwidth}
        \centering
        \includegraphics[width=\textwidth]{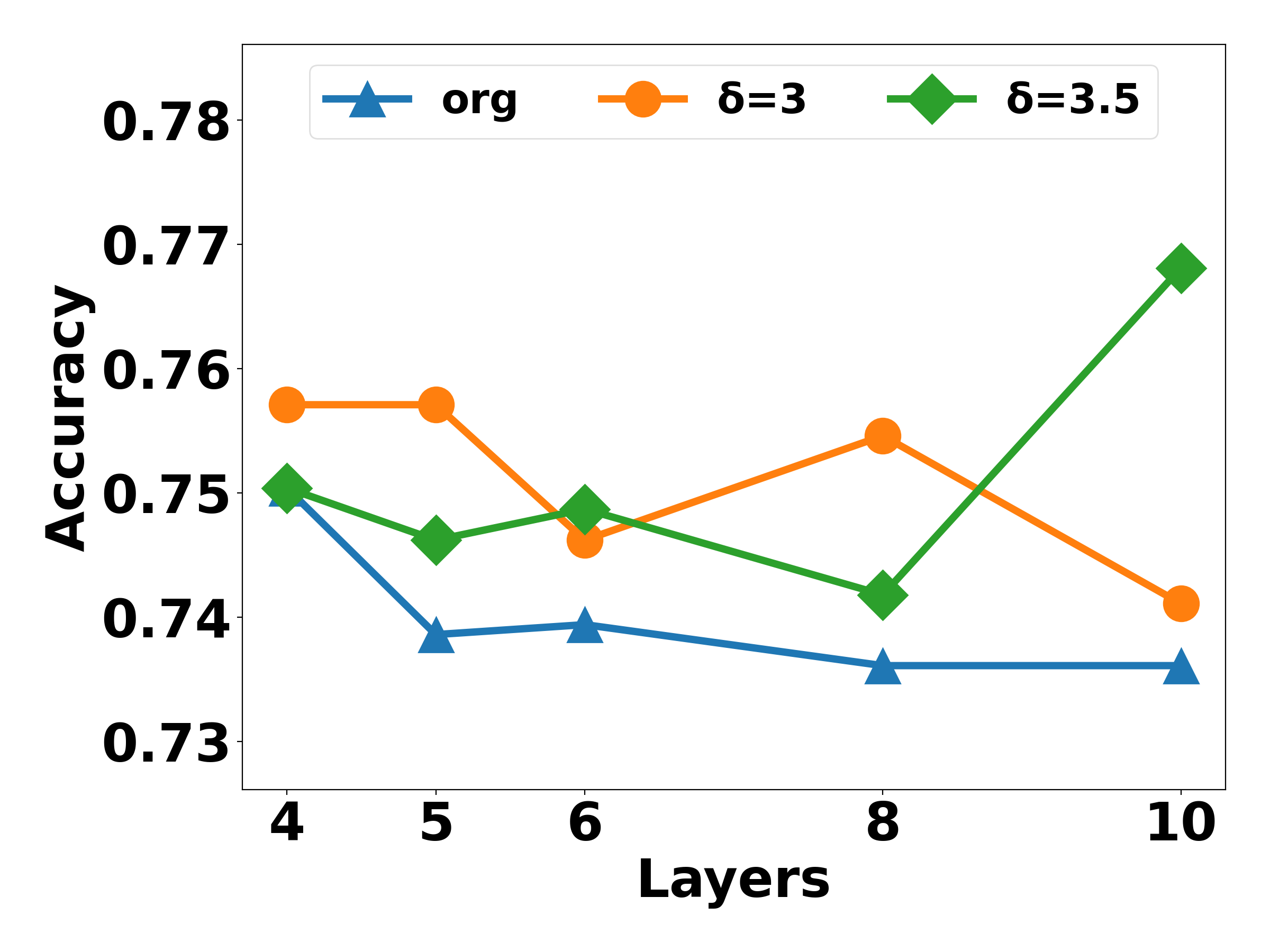}
        \caption{DD(SAGPool)}
        \label{fig:dd_sagpool}
    \end{subfigure}%
    
    \begin{subfigure}[h]{0.24\textwidth}
        \centering
        \includegraphics[width=\textwidth]{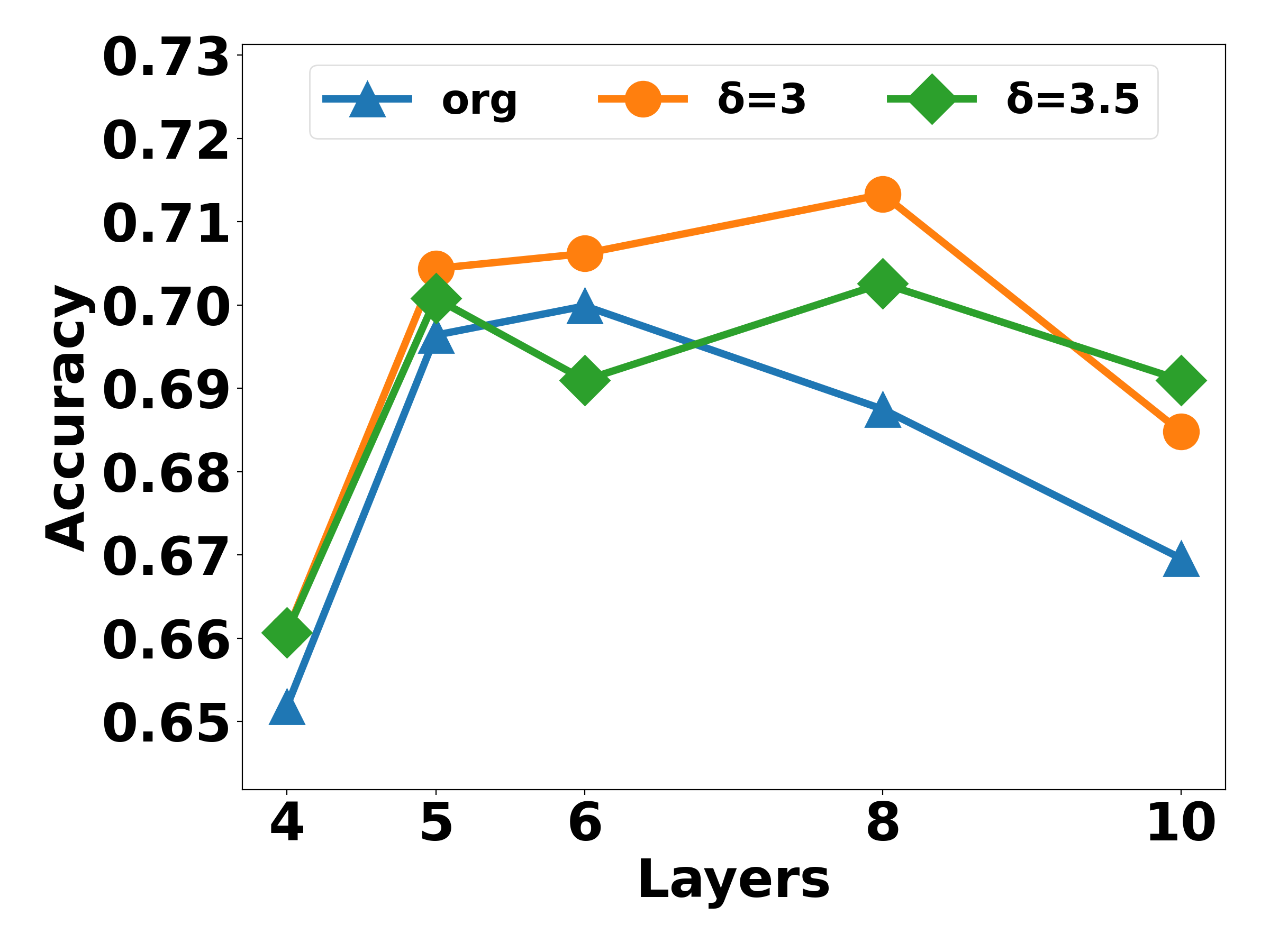}
        \caption{PROT..(GCN)}
        \label{fig:prot_gcn}
    \end{subfigure}%
     ~
    \begin{subfigure}[h]{0.24\textwidth}
        \centering
        \includegraphics[width=\textwidth]{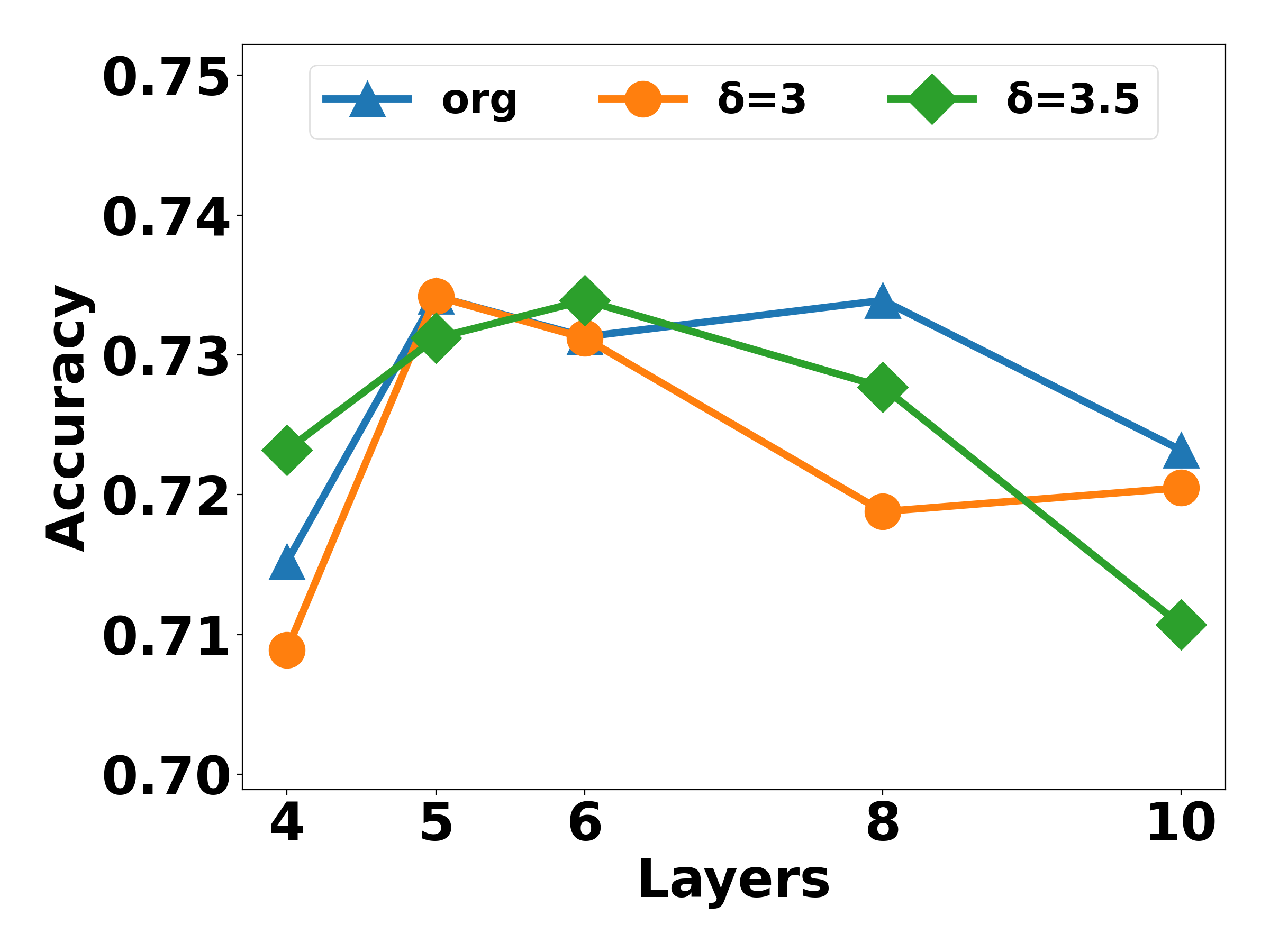}
        \caption{PROT..(GIN-$0$)}
        \label{fig:prot_gin_0}
    \end{subfigure}%
   ~
    \begin{subfigure}[h]{0.24\textwidth}
        \centering
        \includegraphics[width=\textwidth]{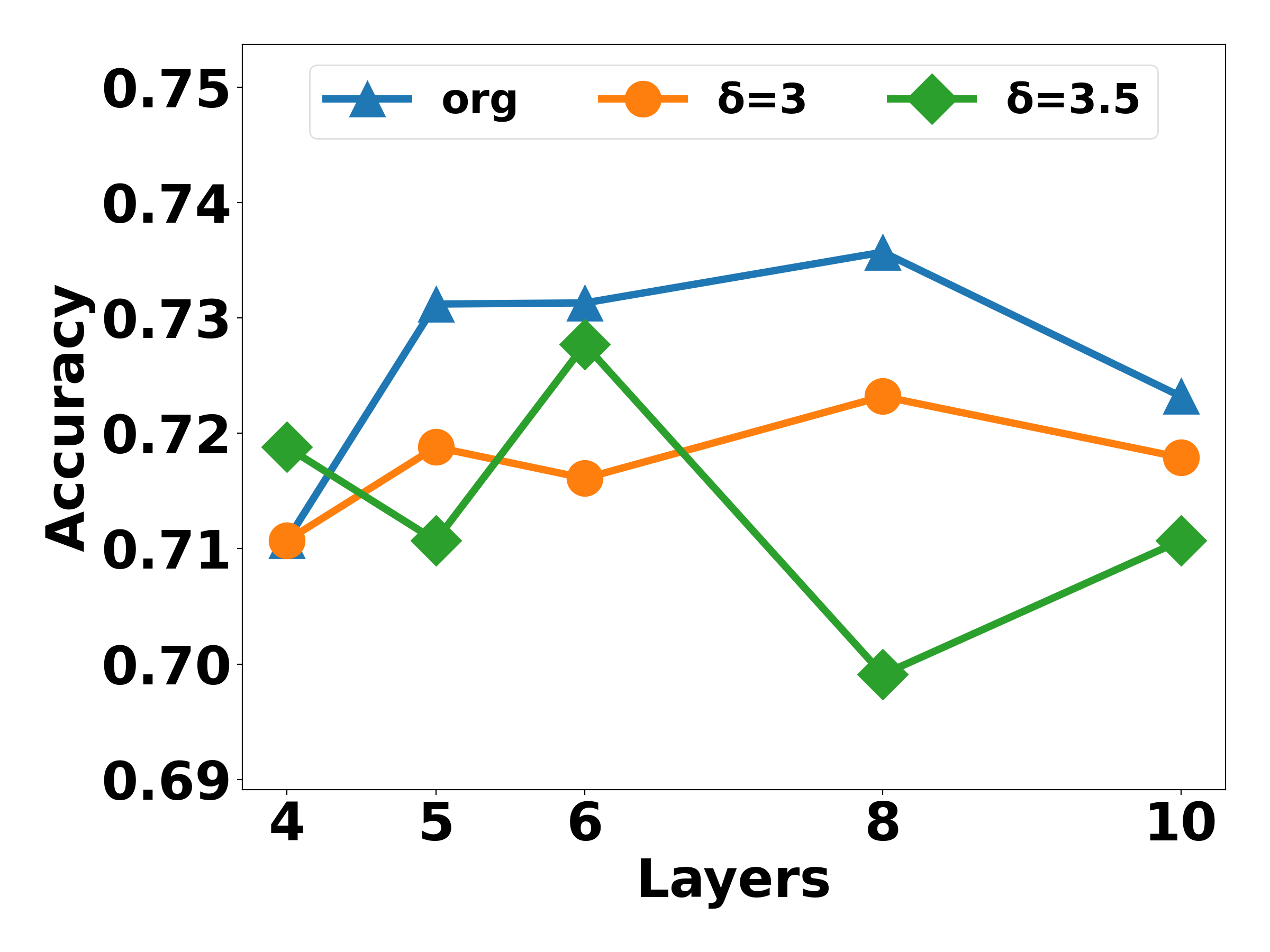}
        \caption{PROT..(GIN-$\epsilon$)}
        \label{fig:prot_gin_e}
    \end{subfigure}%
    ~
    \begin{subfigure}[h]{0.24\textwidth}
        \centering
        \includegraphics[width=\textwidth]{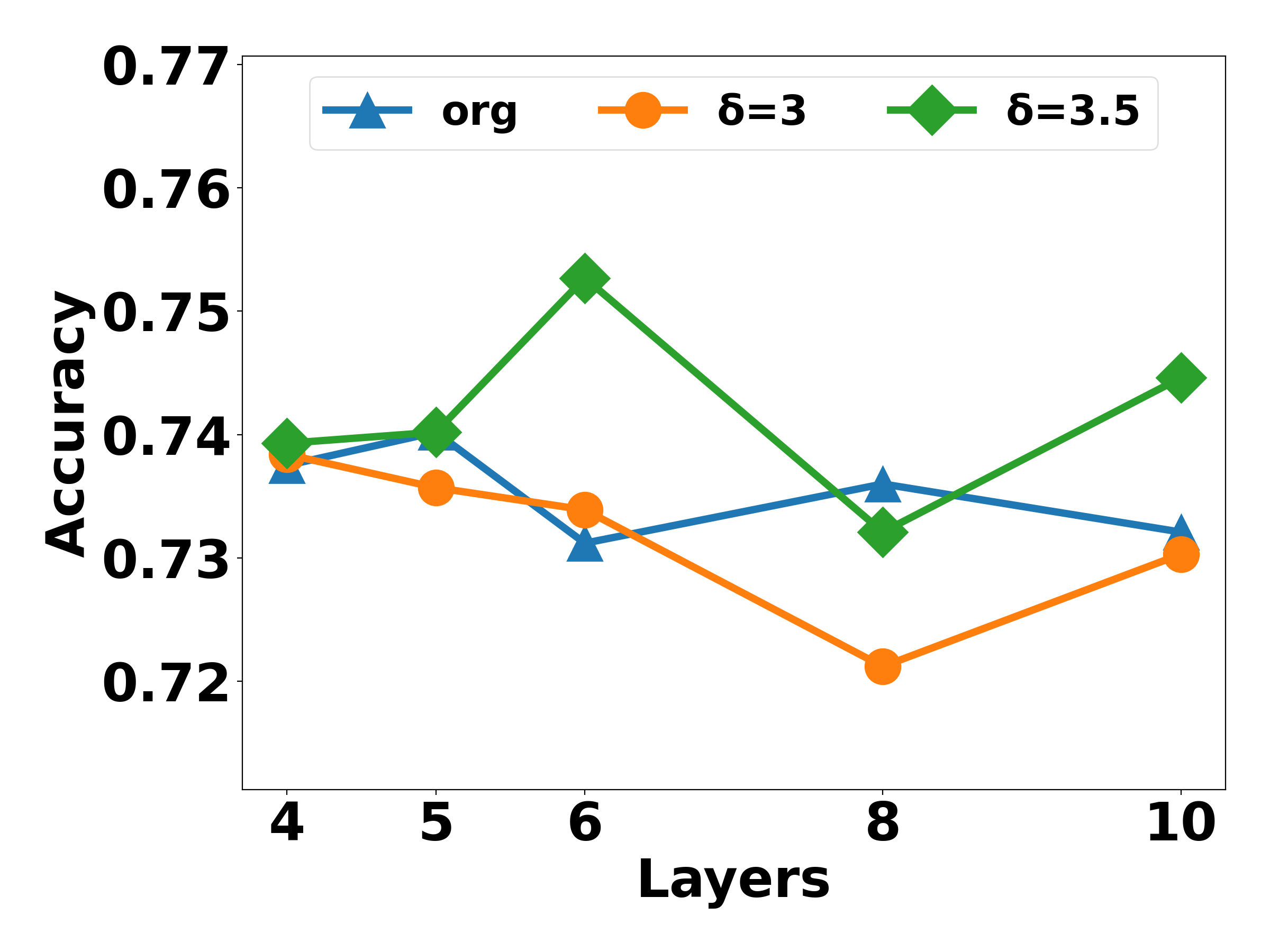}
        \caption{PROT..(SAGPool)}
        \label{fig:prot_sagpool}
    \end{subfigure}%

    \begin{subfigure}[h]{0.24\textwidth}
        \centering
        \includegraphics[width=\textwidth]{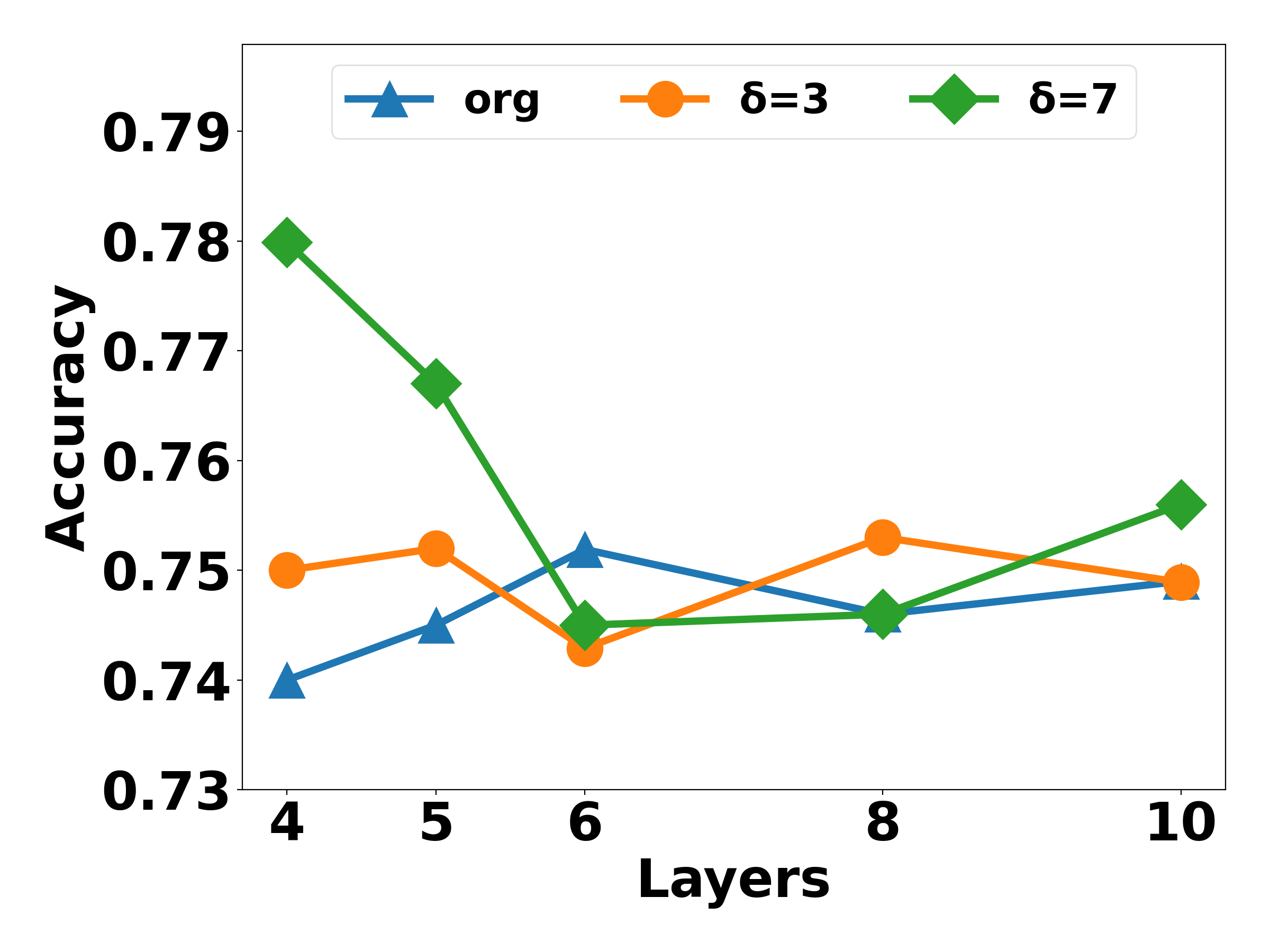}
        \caption{IMDB-B(GCN)}
        \label{fig:imdbb_gcn}
    \end{subfigure}%
     ~
    \begin{subfigure}[h]{0.24\textwidth}
        \centering
        \includegraphics[width=\textwidth]{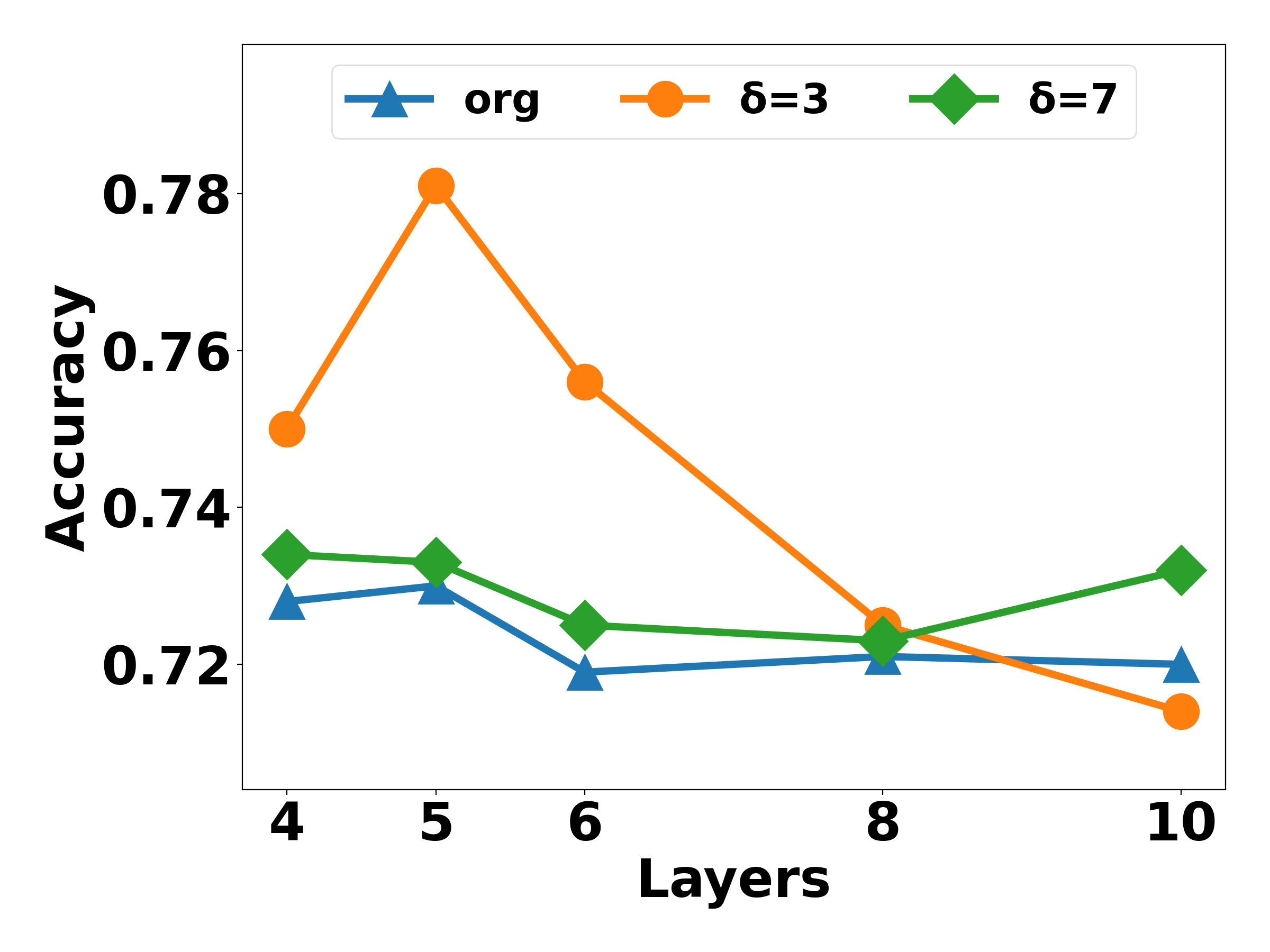}
        \caption{IMDB-B(GIN-$0$)}
        \label{fig:imdbb_gin_0}
    \end{subfigure}%
   ~
    \begin{subfigure}[h]{0.24\textwidth}
        \centering
        \includegraphics[width=\textwidth]{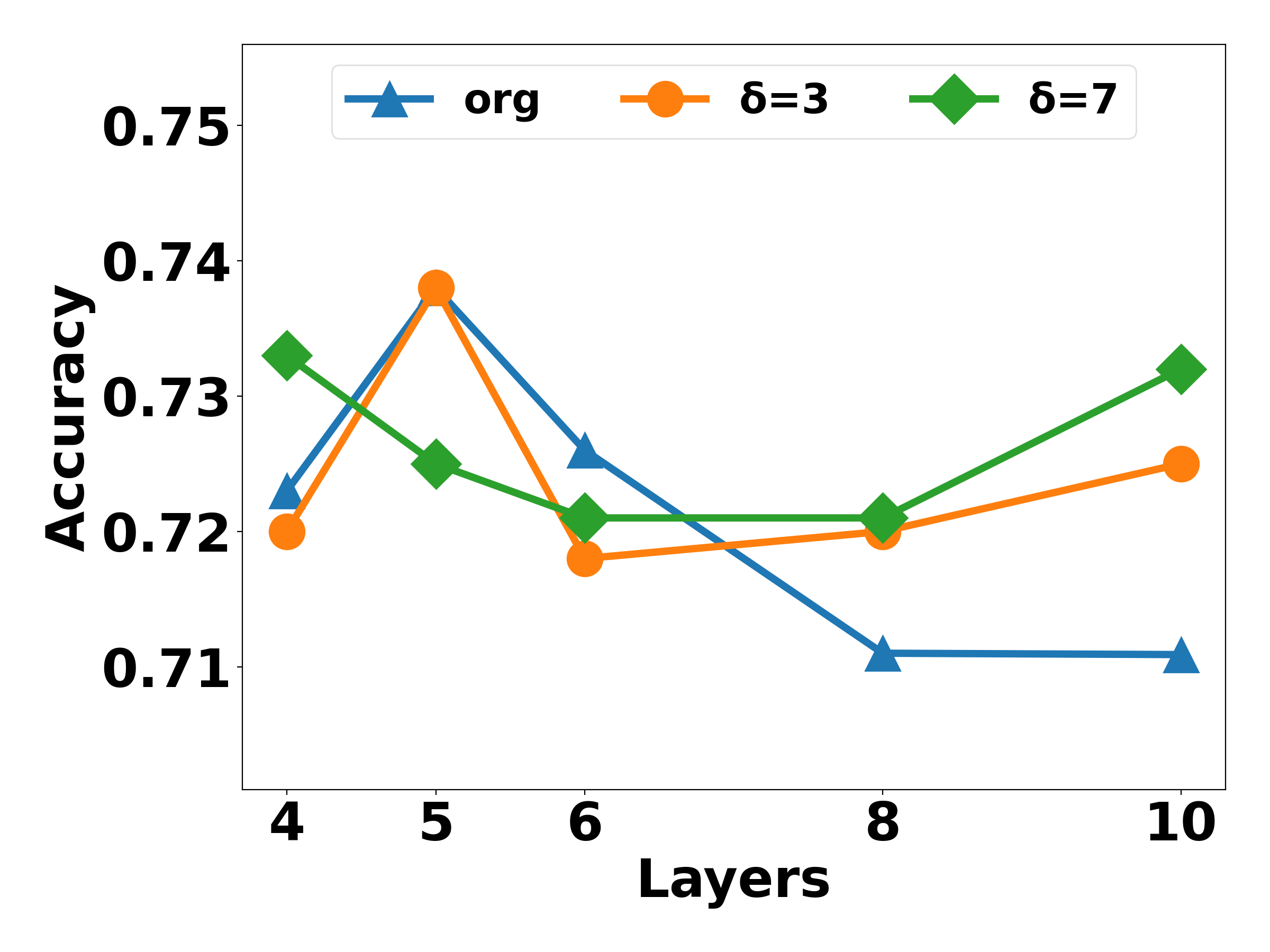}
        \caption{IMDB-B(GIN-$\epsilon$)}
        \label{fig:imdbb_gin_e}
    \end{subfigure}%
    ~
    \begin{subfigure}[h]{0.24\textwidth}
        \centering
        \includegraphics[width=\textwidth]{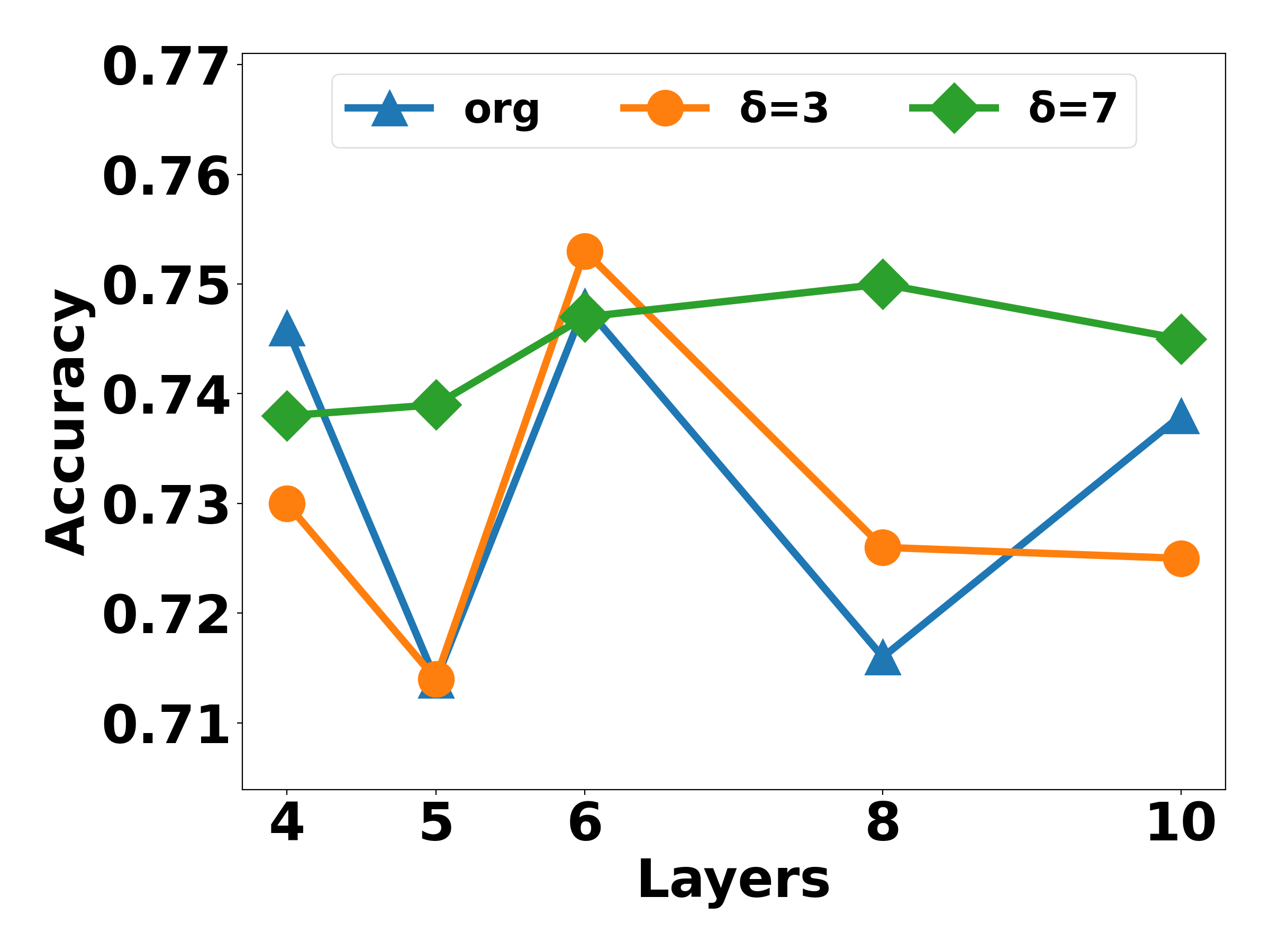}
        \caption{IMDB-B(SAGPool)}
        \label{fig:imdbb_sagpool}
    \end{subfigure}%
    
    \caption{Models' Performance in Deeper Networks}
    \label{fig:layer_exp}
\end{figure*}

\subsection{Analysis in Deeper Network}

We examine the impact of the \TGS~in deeper layers with the backbone models. In this analysis along with the SAGPool, we choose three GNN models: GCN, GIN-$\epsilon$, and GIN-$0$. Besides, we select two datasets from the biomedical domain (DD and PROTEINS) and one from the social network domain (IMDB-BINARY). Figure~\ref{fig:layer_exp} shows the best two ranked (Table~\ref{table:res_details_imdb_red_ptc} and~\ref{table:res_details_dd_prot_nci1_109} in the section~\ref{sec:sup}) \TGS~variants for the threshold $\delta$ compared to the original model performance. Columns 1 (\TGS(GCN)) and 4 (\TGS(SAGPool)) reveal that for increasing the number of layers, in most cases the \TGS~outperforms the original models on all three datasets in multiple layers. Figure~\ref{fig:imdbb_gin_0} and~\ref{fig:imdbb_gin_e} illustrate the similar trends in the context of \TGS(GIN-$\epsilon$) and \TGS(GIN-$0$) on the IMDB-BINARY dataset. However, both of these models show fluctuations in accuracy on the DD and PROTEINS datasets. One interesting fact is that the accuracy trend \TGS(GIN-$0$) has dis-proportionally increased in deeper networks on DD. A possible reason could be the dense nature of the networks in the dataset.

\begin{figure*}[t]
    \centering
    \begin{subfigure}[h]{0.24\textwidth}
        \centering
        \includegraphics[width=\textwidth]{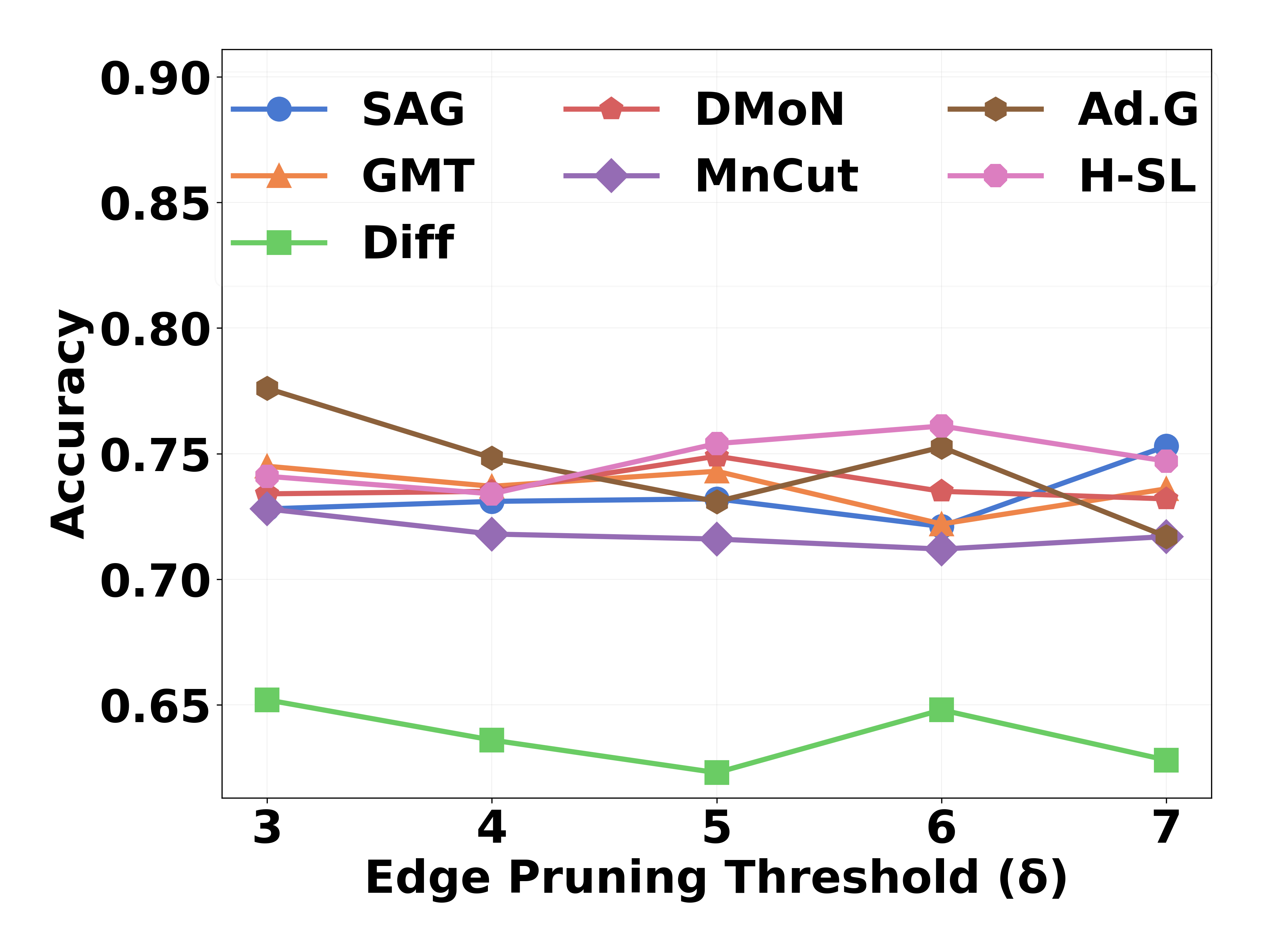}
        \caption{IMDB-B$~\delta$}
        \label{fig: sen_del_ib}
    \end{subfigure}%
     ~
    \begin{subfigure}[h]{0.24\textwidth}
        \centering
        \includegraphics[width=\textwidth]{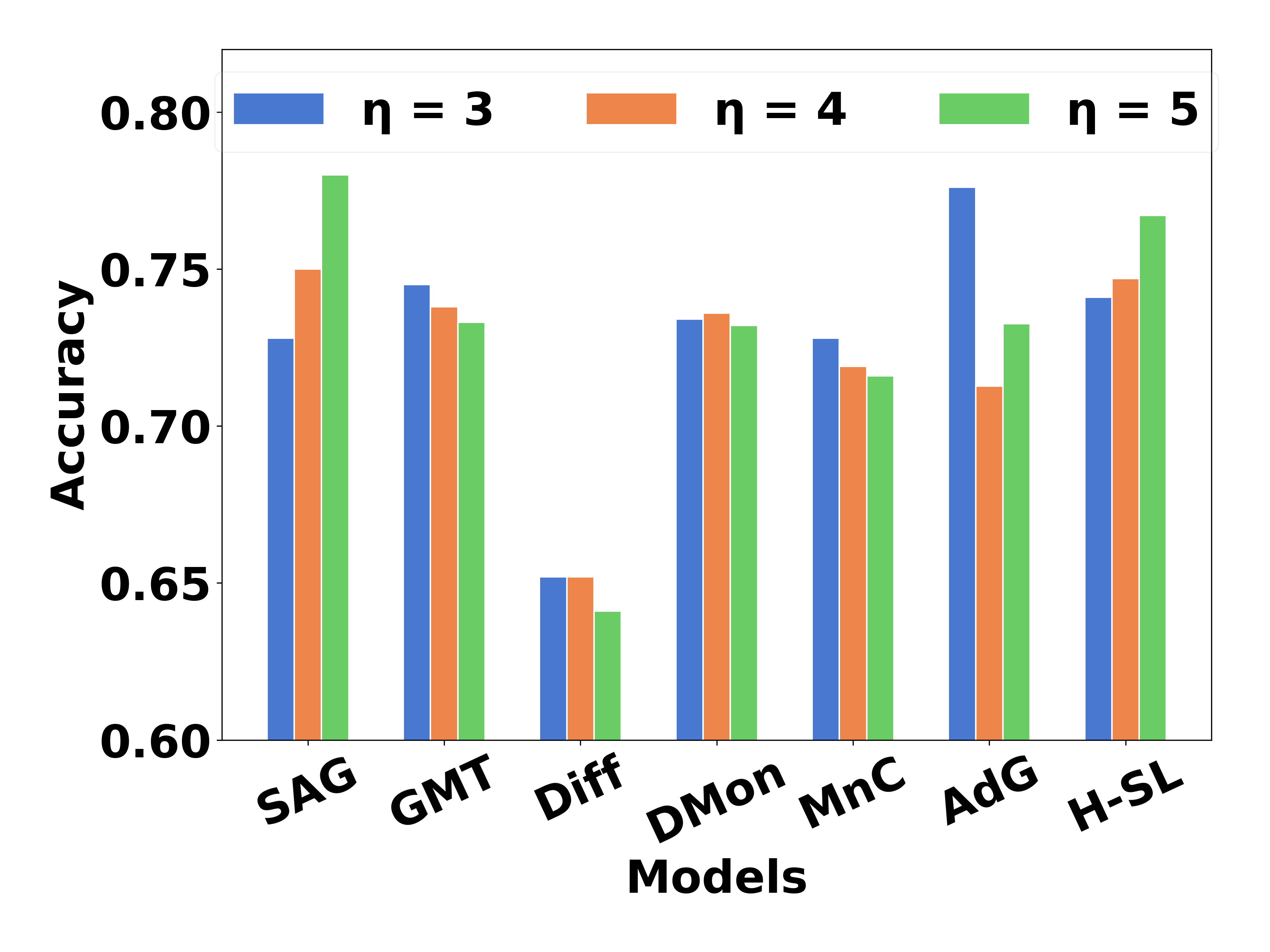}
        \caption{IMDB-B$~\eta$}
        \label{fig: sen_eta_ib}
    \end{subfigure}%
   ~
    \begin{subfigure}[h]{0.24\textwidth}
        \centering
        \includegraphics[width=\textwidth]{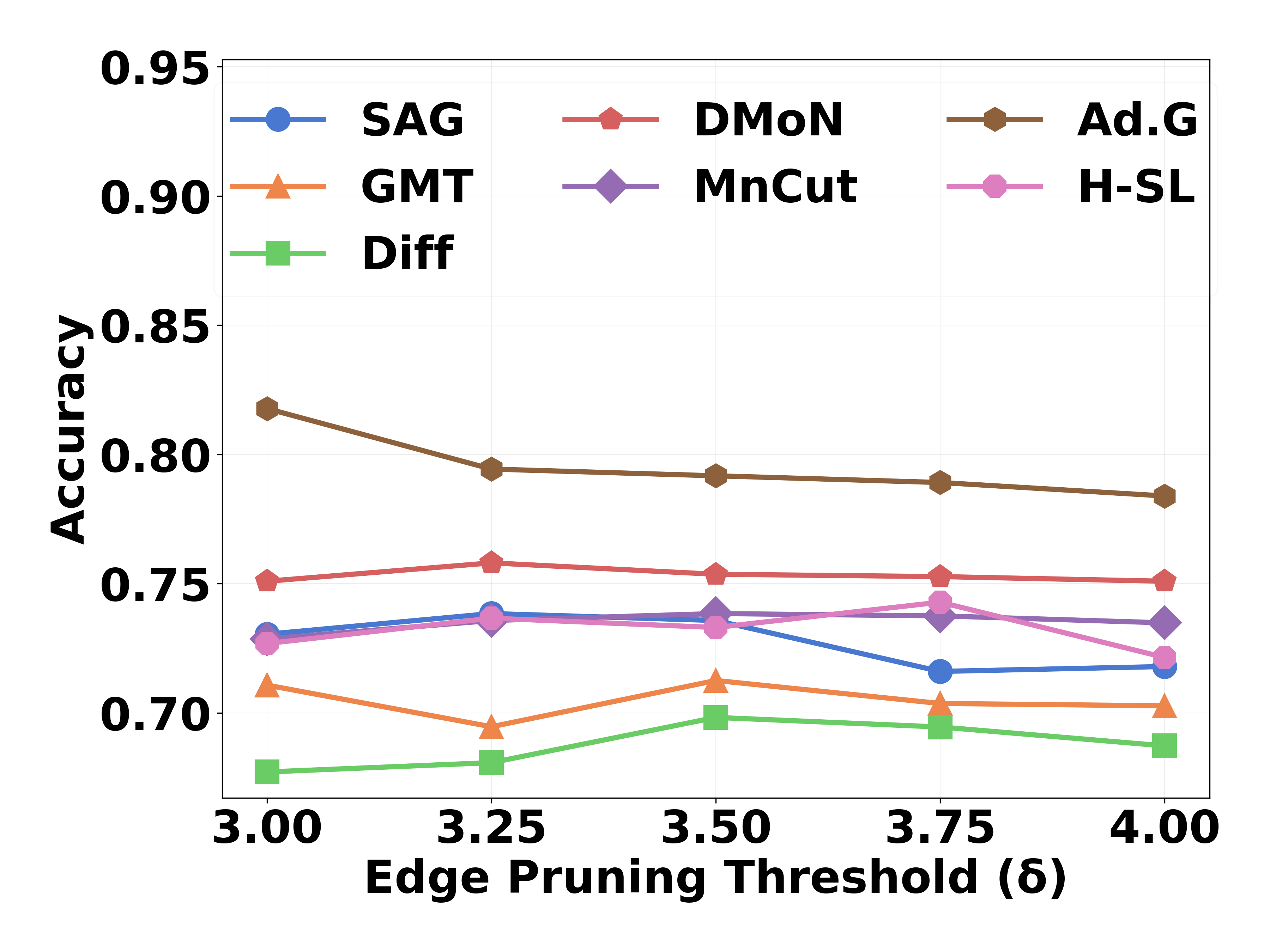}
        \caption{PROTEINS$~\delta$}
        \label{fig: sen_del_pro}
    \end{subfigure}%
    ~
    \begin{subfigure}[h]{0.24\textwidth}
        \centering
        \includegraphics[width=\textwidth]{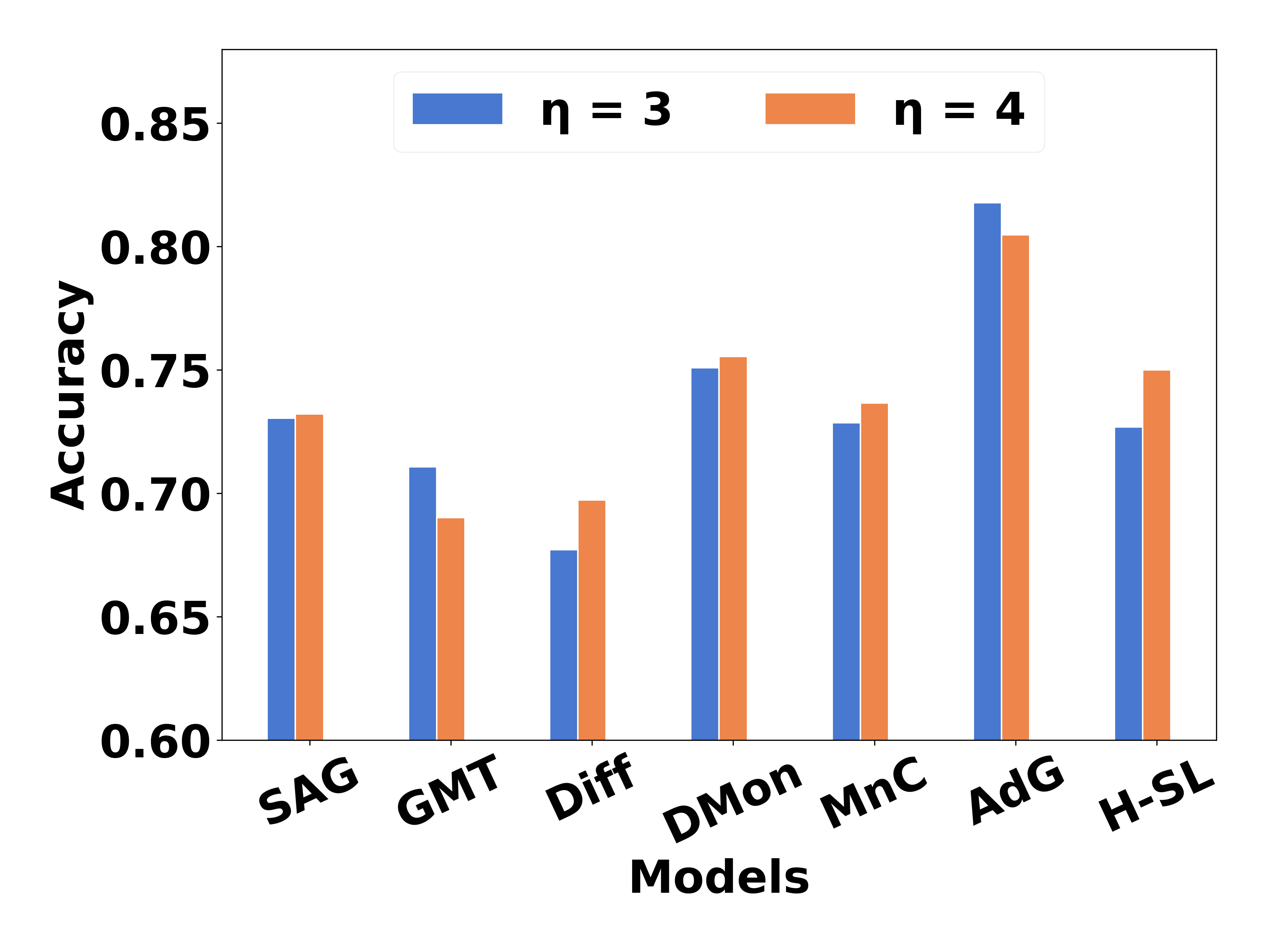}
        \caption{PROTEINS$~\eta$}
        \label{fig: sen_eta_pro}
    \end{subfigure}%

    \caption{Change of parameters $\delta$ and $\eta$ on IMDB-BINARY and PROTEINS.}
    \label{fig:sesitivity}
\end{figure*}

\subsection{Sensitivity Analysis}
In Figure~\ref{fig:sesitivity}, we demonstrate our model’s performance for variations of hyperparameters’ values on the IMDB-BINARY and PROTEINS datasets. Notably, in most cases, the pruning rate decreases as much as the delta value increases. Figure~\ref{fig: sen_del_ib}~shows at $\delta = 3$ value, our equipped \TGS~models perform well. We observe that for lower $\delta$ value, the accuracy of \TGS~with AdamGNN increases, whereas degrades for \TGS(SAGPool). On the other hand, in the PROTEINS dataset (Figure~\ref{fig: sen_del_pro}), for changing the $\delta$ value, \TGS~decorated with SAGPool, MinCutPool, DMonPool, and HGP-SL, showing near-consistent performance. However, when increasing the $\delta$ value from $3$ to $3.25$, \TGS(AdamGNN)’s accuracy decreases and shows an almost stable performance. Assembled with GMT and Diffpool, the accuracy of \TGS~shows some variations from $\delta = (3-3.5)$ and then remains near the same score for other values.  

Regarding the change of cutoff variable $\eta$, figures~\ref{fig: sen_eta_ib}~and~\ref{fig: sen_eta_pro} show the performance changes on the same datasets. Similar to $\delta$, the number of pruned edges in the graph decreases for increasing the value $\eta$, and \TGS(SAGPool)’s accuracy increases. In contrast, for the same reason, the performance of our skill with backbone AdamGNN and HGP-SL degrades. Other models display minor fluctuations in accuracy with the change of $\eta$ value.
% In Figure~\ref{fig:sesitivity}, we demonstrate our equipped \TGS~model’s performance for variations of hyperparameters value on the IMDB-B and PROTEINS datasets. Generally, in most cases, the pruning rate decreases for increasing the $\delta$. Both line graphs for $(\delta~vs~ Accuracy)$ signify that the pruning of high-truss edges does not lose the core graph structure, as in most cases, our \TGS~with backbones attain a similar or better performance. So, those dense edges have little contribution to graph structure learning with GNN and graph pooling models. 

\begin{table*}[h!]
\centering
\caption{Ablation Study Table. Boldly marked entries perform better than \TGS.}
\label{table: abl_study_table}
\begin{tabular}{|c|c|cccccc|}
\hline
 \multicolumn{2}{|c|}{ \textbf{Changes}} & SAGPool & GMT & DMonPool & MnCutPool & DiffPool & AdamGNN \\ \hline
\multirow{4}{*}{\cellcolor{gray!15} 
 \textbf{IMDB-BINARY}} & \cellcolor{gray!15} avg (avg, avg) & 75.00 & \textbf{74.59} & 74.70 & \textbf{73.09} & 63.00 & 75.43 \\ 
 & \cellcolor{gray!15} min (min, min) & 74.00 & 48.06 & 72.70 & \textbf{73.13} & \textbf{65.10} & \textbf{79.77} \\ 
 & \cellcolor{gray!15} Prune 2* & 70.99 & 48.69 & 51.80 & 48.50 & 61.10 & 53.73 \\ 
 \cellcolor{gray!15}  & \cellcolor{gray!15} Prune 3* & 68.70 & 47.90 & 54.89 & 49.59 & 62.79 & 59.80 \\ \hline
\multirow{4}{*}{\cellcolor{gray!15} \textbf{IMDB-MULTI}} & \cellcolor{gray!15} avg (avg, avg) & 44.67 & 47.40 & 48.86 & 50.59 & 41.46 & 50.04 \\ 
& \cellcolor{gray!15} min (min, min) & 46.67 & \textbf{48.06} & 49.13 & 50.73 & 38.86 & 46.92 \\ 
& \cellcolor{gray!15} Prune 2* & 43.99 & 47.93 & 49.73 & 49.09 & 39.47 & 50.04 \\ 
\cellcolor{gray!15}  & \cellcolor{gray!15} Prune 3* & \textbf{50.59} & 47.46 & 49.40 & 50.99 & 41.06 & 47.96 \\ \hline
\multirow{4}{*}{\cellcolor{gray!15} \textbf{PROTEINS}} & \cellcolor{gray!15} avg (avg, avg) & 71.14 & 70.62 & \textbf{75.53} & \textbf{74.01} & \textbf{69.99} & 81.51 \\ 
& \cellcolor{gray!15} min (min, min) & 69.64 & 67.41 & 73.83 & \textbf{74.10} & \textbf{71.01} & 74.10 \\ 
& \cellcolor{gray!15} Prune 2* & 72.40 & 66.78 & 73.84 & \textbf{74.37} & \textbf{70.27} & 69.79 \\ 
\cellcolor{gray!15}  & \cellcolor{gray!15} Prune 3* & 72.58 & 66.60 & 59.46 & 62.05 & 69.54 & 71.35 \\ \hline
\multirow{4}{*}{\cellcolor{gray!15} \textbf{NCI1}} & \cellcolor{gray!15} avg (avg, avg) & 70.07 & 61.55 & 68.8 & 69.97 & 68.10 & 47.36 \\ 
& \cellcolor{gray!15} min (min, min) & \textbf{73.47} & 61.19 & 69.05 & 73.52 & 68.17 & 61.22 \\ 
& \cellcolor{gray!15} Prune 2* & 67.73 & 60.97 & 69.63 & 70.34 & 68.17 & 58.16 \\ 
\cellcolor{gray!15}  & \cellcolor{gray!15} Prune 3* & 68.71 & 59.17 & 68.83 & 70.07 & 67.15 & 60.02 \\ \hline
\end{tabular}
\end{table*}

\subsection{Ablation Study} 
This section observes the strategical and functional significance of \TGS~with the backbone graph pooling models. We chose four datasets and six pooling methods. In Table~\ref{table: abl_study_table}, for each dataset at the first two rows, we change the edge's connectivity strength measuring equations~(\ref{eq:6}), and (\ref{eq:7}). In the first row, the equation~(\ref{eq:6}) remains the same but at equation~(\ref{eq:7}) instead of $minimum$ the $average$ of two end nodes' strength has been taken. On the other hand, in the second row, node strength measuring equation~(\ref{eq:6}) is modified whereas the other equation remains unchanged. In the last two rows, we change the pruning procedure, examining to prune two (prune~$2^{*}$) and three (prune~$3^{*}$) edges without updating the edge trussness. Due to the change of equations, the model's performance on the PROTEINS dataset increases with its extension to MinCutPool and DiffPool methods. However, on the NCI1 dataset, the performance of AdamGNN dramatically decreases (47.36\% and 61.22\%).
Regarding examining 2 and 3 edges for pruning, sometimes more than one edge is pruned without updating the other edges' trussness. Hence, significant information processing connections could prune in the system. Compared to the result in Table~\ref{table:tbl_result}, our model's performance severely degrades on some datasets with the components change. Nonetheless, the modified pruning strategy with MinCutPool and DiffPool achieve better results over \TGS~are 74.37\%  and 70.27\%  respectively on the PROTEINS dataset.

\section{Conclusion}
In this paper, we have proposed an effective k-truss-based graph sparsification model to facilitate graph learning of the graph neural networks (GNN). Through the sparsification of dense graph regions' overflowed message passing edges, our model includes more variability to the input graph for alleviating oversmoothing. Comprehensive experiments on eight renowned datasets verify that \TGS~is consistent in performance over popular graph pooling and readout-based GNN models. We expect our research to show some interesting directions: Learning the edge pruning threshold during training, applying parallelization during pruning edges at different k-truss subgraphs, and joint learning to measure edge importance during graph sparsification.

\bibliographystyle{abbrv}
\bibliography{TGS_Arxiv.bib}

\section{Supplement}\label{sec:sup}

\subsection{Result Details}
This section reports the experiment details of the \TGS~model pipelined with the backbone graph pooling and GNN models. In all the tables, the social networks and biomedical domains' datasets' results are shown with the accuracy ($\%$) metric. We measure the average accuracy for each dataset and rank them for different variations of the edge pruning threshold $\delta$ compared to the original backbone model's scores. Table~\ref{table:res_details_imdb_red_ptc} and~\ref{table:res_details_dd_prot_nci1_109} report all the results of different \TGS-variants for separate $\delta$ values. Due to limited space, we represent the (REDDIT-BINARY \& PTC)  and (NCI1 \& NCI109) datasets' results together in sub-tables. 

% Table 1: IMDB-BINARY
\begin{table}[h!]

\caption{IMDB-B, IMDB-M, REDDIT-B and PTC results}
% \scriptsize

\centering
\begin{tabular}{|c|c|c|c|c|c|c|c|}
\hline
\multicolumn{7}{|c|}{\cellcolor{gray!15} \textbf{IMDB-BINARY}} \\ \hline 
\multicolumn{1}{|c|}{} & \multicolumn{1}{c|}{} & \multicolumn{5}{c|}{$\bm{\delta}$} \\ \cline{3-7}
 & \textbf{Original} & $\mathbf{3}$ & $\mathbf{4}$ & $\mathbf{5}$ & $\mathbf{6}$ & $\mathbf{7}$ \\ \hline
SAGPool & 72.80 & 73.10 & 73.10 & 73.20 & 72.10 & 75.30 \\ \hline
GMT & 71.10 & 74.50 & 73.70 & 72.20 & 72.20 & 73.60 \\ \hline
DiffPool & 63.90 & 65.20 & 63.60 & 62.30 & 64.80 & 62.80 \\ \hline
DMon & 74.00 & 73.40 & 73.50 & 74.90 & 73.50 & 73.20 \\ \hline
MinCut & 71.40 & 72.80 & 71.80 & 71.60 & 71.20 & 71.70 \\ \hline
AdamGNN & 72.48 & 77.60 & 74.83 & 73.09 & 75.26 & 71.70 \\ \hline
HGP-SL & 72.90 & 74.10 & 73.40 & 75.40 & 76.10 & 74.70 \\ \hline
GIN-$\epsilon$ & 73.80 & 73.80 & 73.90 & 73.00 & 73.40 & 72.50 \\ \hline
GIN-$0$ & 73.00 & 78.10 & 69.50 & 71.60 & 73.20 & 73.30 \\ \hline
GCN & 74.00 & 75.00 & 79.00 & 78.99 & 74.00 & 77.99 \\ \hline \hline
\cellcolor{gray!15} Mean & 71.94 & 73.76 & 72.63 & 72.62 & 72.58 & 72.68 \\ \hline
\cellcolor{gray!15} Rank & 6 & 1 & 3 & 4 & 5 & 2 \\ \hline
\end{tabular}

\vspace{0.5em}

\centering
\begin{tabular}{|c|c|c|c|c|c|c|c|}
\hline
\multicolumn{7}{|c|}{\cellcolor{gray!15} \textbf{IMDB-MULTI}} \\ \hline
\multicolumn{1}{|c|}{} & \multicolumn{1}{c|}{} & \multicolumn{5}{c|}{$\bm{\delta}$} \\ \cline{3-7}
& \textbf{Original} & $\mathbf{3}$ & $\mathbf{4}$ & $\mathbf{5}$ & $\mathbf{6}$ & $\mathbf{7}$ \\ \hline
SAGPool & 44.86 & 42.67 & 44.67 & 41.47 & 42.73 & 43.00 \\ \hline
GMT & 47.40 & 47.60 & 45.53 & 46.27 & 47.93 & 47.13 \\ \hline
DiffPool & 44.80 & 41.53 & 40.93 & 40.67 & 40.67 & 42.67 \\ \hline
DMon & 49.53 & 49.30 & 48.4 & 48.46 & 49.60 & 48.33 \\ \hline
MinCut & 51.06 & 50.06 & 51.20 & 50.73 & 50.20 & 50.93 \\ \hline
AdamGNN & 49.53 & 46.92 & 46.40 & 50.57 & 50.05 & 46.40 \\ \hline
HGP-SL & 49.46 & 49.80 & 48.80 & 48.87 & 47.33 & 49.00 \\ \hline
GIN-$\epsilon$ & 49.33 & 43.80 & 49.00 & 47.93 & 50.80 & 53.40 \\ \hline
GIN-$0$ & 47.60 & 41.47 & 46.13 & 48.73 & 51.00 & 52.53 \\ \hline
GCN & 41.33 & 44.67 & 40.67 & 42.67 & 46.67 & 40.67 \\ \hline \hline
\cellcolor{gray!15} Mean & 47.49 & 45.78 & 46.17 & 46.64 & 47.70 & 47.41 \\ \hline
\cellcolor{gray!15} Rank & 2 & 6 & 5 & 4 & 1 & 3 \\ \hline

\end{tabular}
% \end{table}

% \begin{table}[t]
% \caption{REDDIT-BINARY and PTC Results}

\vspace{0.5em}

\centering
\begin{tabular}{|c|c|c|c|c|c|c|c|} 
\hline
& \multicolumn{4}{|c|}{\cellcolor{gray!15} \textbf{REDDIT-BINARY}} & \multicolumn{2}{|c|}{\cellcolor{gray!15} \textbf{PTC}} \\ \cline{2-7}
\multicolumn{1}{|c|}{} & \multicolumn{1}{c|}{} & \multicolumn{3}{c|}{$\bm{\delta}$} & \multicolumn{1}{c|}{} & \multicolumn{1}{c|}{$\bm{\delta}$} \\ \cline{3-5} \cline{7-7}
& \textbf{Org.} & $\mathbf{3}$ & $\mathbf{3.5}$ & $\mathbf{4}$ & \textbf{Org.} & $\mathbf{2.5}$ \\ \hline
% & \textbf{Org.} & $\mathbf{\delta=3}$ & $\mathbf{\delta=3.5}$ & $\mathbf{\delta=4}$ & \textbf{Org.} & $\mathbf{\delta=2.5}$ \\ \hline
SAGPool & 77.55 & 76.15 & 79.05 & 77.80 & 57.14 & 59.14 \\ \hline
GMT & 70.95 & 70.95 & 71.05 & 71.25 & 50.86 & 51.14 \\ \hline
DiffPool & 80.72 & 81.45 & 82.32 & 82.03 & 46.29 & 55.14 \\ \hline
DMon & 84.95 & 85.75 & 84.70 & 85.85 & 53.71 & 56.00 \\ \hline
MinCut & 76.85 & 77.05 & 76.19 & 76.45 & 56.28 & 58.57 \\ \hline
AdamGNN & OOM & OOM & OOM & OOM & 60.00 & 62.86 \\ \hline
HGP-SL & OOM & OOM & OOM & OOM & 56.00 & 57.43 \\ \hline
GIN-$\epsilon$ & 73.55 & 73.75 & 74.00 & 74.55 & 68.28 & 66.57 \\ \hline
GIN-$0$ & 73.60 & 73.65 & 74.10 & 73.30 & 68.86 & 65.43 \\ \hline
GCN & 88.99 & 89.50 & 88.50 & 73.50 & 45.71 & 45.71 \\ \hline \hline
\cellcolor{gray!15} Mean & 78.40 & 78.53 & 78.74 & 76.84 & 56.31 & 57.80 \\ \hline
\cellcolor{gray!15} Rank & 3 & 2 & 1 & 4 & 2 & 1 \\ \hline
\end{tabular}
\label{table:res_details_imdb_red_ptc}
\end{table}

\begin{table}[h!]
\caption{ DD, PROTEINS, NCI1 and NCI109 results}
% \scriptsize
\centering
\begin{tabular}{|c|c|c|c|c|c|c|c|}
\hline
\multicolumn{7}{|c|}{\cellcolor{gray!15} 
 \textbf{DD}} \\ \hline
\multicolumn{1}{|c|}{} & \multicolumn{1}{c|}{} & \multicolumn{5}{c|}{$\bm{\delta}$} \\ \cline{3-7}
& \textbf{Original} & $\mathbf{3}$ & $\mathbf{3.25}$ & $\mathbf{3.5}$ & $\mathbf{3.75}$ & $\mathbf{4}$ \\ \hline
SAGPool & 77.31 & 79.83 & 78.15 & 80.67 & 80.67 & 79.00 \\ \hline
GMT & 65.88 & 69.5 & 67.39 & 68.49 & 64.62 & 63.78 \\ \hline
DiffPool & 75.31 & 78.13 & 78.75 & 78.13 & 77.81 & 73.75 \\ \hline
DMon & 79.32 & 80.42 & 79.49 & 79.57 & 79.66 & 80.08 \\ \hline
MinCut & 76.63 & 77.31 & 77.22 & 76.97 & 76.55 & 76.72 \\ \hline
AdamGNN & 64.63 & 71.65 & 74.25 & 71.12 & 67.87 & 71.00 \\ \hline
HGP-SL & 72.35 & 71.68 & 72.44 & 71.43 & 72.86 & 73.19 \\ \hline
GIN-$\epsilon$ & 76.44 & 73.73 & 71.36 & 74.58 & 73.9 & 74.75 \\ \hline
GIN-$0$ & 74.58 & 73.22 & 74.75 & 75.08 & 74.49 & 75.93 \\ \hline
GCN & 58.82 & 73.95 & 66.38 & 67.23 & 70.58 & 69.75 \\ \hline \hline
\cellcolor{gray!15} Mean & 72.13 & 74.94 & 74.02 & 74.33 & 73.90 & 73.80 \\ \hline
\cellcolor{gray!15} Rank & 6 & 1 & 3 & 2 & 4 & 5 \\ \hline
\end{tabular}

\vspace{0.5em}

\centering
\begin{tabular}{|c|c|c|c|c|c|c|c|}
\hline
\multicolumn{7}{|c|}{\cellcolor{gray!15} 
 \textbf{PROTEINS}} \\ \hline
\multicolumn{1}{|c|}{} & \multicolumn{1}{c|}{} & \multicolumn{5}{c|}{$\bm{\delta}$} \\ \cline{3-7}
& \textbf{Original} & $\mathbf{3}$ & $\mathbf{3.25}$ & $\mathbf{3.5}$ & $\mathbf{3.75}$ & $\mathbf{4}$ \\ \hline
SAGPool & 72.68 & 73.04 & 73.84 & 73.57 & 71.60 & 71.79 \\ \hline
GMT & 70.63 & 71.07 & 69.46 & 71.25 & 70.36 & 70.27 \\ \hline
DiffPool & 71.10 & 67.71 & 68.07 & 69.82 & 69.45 & 68.72 \\ \hline
DMon & 75.45 & 75.09 & 75.8 & 75.36 & 75.27 & 75.09 \\ \hline
MinCut & 73.75 & 72.86 & 73.57 & 73.84 & 73.75 & 73.48 \\ \hline
AdamGNN & 78.12 & 81.77 & 79.43 & 79.17 & 78.91 & 78.39 \\ \hline
HGP-SL & 74.64 & 72.68 & 73.66 & 73.30 & 74.28 & 72.14 \\ \hline
GIN-$\epsilon$ & 73.12 & 71.88 & 71.88 & 71.07 & 73.39 & 70.57 \\ \hline
GIN-$0$ & 73.42 & 73.42 & 73.84 & 73.12 & 72.78 & 72.50 \\ \hline
GCN & 65.18 & 66.07 & 63.39 & 66.07 & 64.28 & 63.39 \\ \hline \hline
\cellcolor{gray!15} Mean & 72.81 & 72.56 & 72.29 & 72.66 & 72.41 & 71.63 \\ \hline
\cellcolor{gray!15} Rank & 1 & 3 & 5 & 2 & 4 & 6 \\ \hline
\end{tabular}

\vspace{0.5em}

\centering
\begin{tabular}{|c|c|c|c|c|c|c|c|}
\hline
& \multicolumn{3}{|c|}{\cellcolor{gray!15}  \textbf{NCI1}} & \multicolumn{3}{|c|}{\cellcolor{gray!15} 
 \textbf{NCI109}} \\ \cline{2-7} 
\multicolumn{1}{|c|}{} & \multicolumn{1}{c|}{} & \multicolumn{2}{c|}{$\bm{\delta}$} & \multicolumn{1}{c|}{} & \multicolumn{2}{c|}{$\bm{\delta}$} \\ \cline{3-4} \cline{6-7}
& \textbf{Org.} & $\mathbf{2.5}$ & $\mathbf{3}$ & \textbf{Org.} & $\mathbf{2.5}$ & $\mathbf{3}$ \\ \hline
SAGPool & 70.10 & 68.49 & 70.72 & 67.37 & 66.84 & 67.07  \\ \hline
GMT & 60.29 & 57.42 & 58.44 & 48.86 & 51.01 & 51.13 \\ \hline
DiffPool & 66.96 & 67.64 & 68.18 & 67.60 & 65.97 & 67.2  \\ \hline
DMon & 74.20 & 74.23 & 74.01 & 72.61 & 73.50 & 73.38  \\ \hline
MinCut & 81.85 & 82.46 & 82.50 & 73.28 & 73.04 & 72.56  \\ \hline
AdamGNN & 81.70 & 81.21 & 81.87 & 65.16 & 61.79 & 67.81 \\ \hline
HGP-SL & 71.53 & 73.48 & 70.07 & 72.87 & 72.58 & 72.56  \\ \hline
GIN-$\epsilon$ & 81.85 & 82.46 & 82.50 & 75.93 & 75.98 & 76.61  \\ \hline
GIN-$0$ & 81.70 & 81.21 & 81.87 & 74.99 & 74.53 & 74.94  \\ \hline
GCN & 71.53 & 73.48 & 70.07 & 75.60 & 76.81 & 72.70  \\ \hline \hline
\cellcolor{gray!15} Mean & 74.17 & 74.21 & 74.02 & 69.43 & 69.21 & 69.60 \\ \hline
\cellcolor{gray!15} Rank & 2 & 1 & 3 & 2 & 3 & 1  \\ \hline
\end{tabular}
\label{table:res_details_dd_prot_nci1_109}
\end{table}

 % Force the table to be processed
\begin{table}[t!]
\caption{Training Parameters in Models (Part 1)}
\centering
\begin{tabular}{|c|c|c|c|c|}
\hline
\cellcolor{gray!15} \textbf{model} & \cellcolor{gray!15} \textbf{lr. rate} & \cellcolor{gray!15} \textbf{\# epochs} & \cellcolor{gray!15} \textbf{\# layers} & \cellcolor{gray!15} \textbf{hid. size} \\ \hline
SAGPool  & 0.005 & 1,000,000 & 3 & 128 \\ \hline
GMT & 0.0005 & 500 & 3 & 32 \\ \hline
DiffPool & 0.001 & 500 & 3 & 64 \\ \hline
DMon & 0.001 & 500 & 3 & 32 \\ \hline
MinCut & 0.0005 & 15,000 & 3 & 32 \\ \hline
AdamGNN & 0.01 & 200 & 1 or 2 & 64 \\ \hline
HGP-SL & 0.001 & 1,000 & 3 & 128 \\ \hline
GIN-($\epsilon$ \& $0$) & 0.01 & 350 & 5 & 16 \\ \hline
GCN & 0.005 & 350 & 4 & 128 \\ \hline
\end{tabular}
\label{table:param_details_1}
\end{table}

\begin{table}[t!]
\caption{Training Parameters in Models (Part 2)}
\centering
\begin{tabular}{|c|c|c|c|c|}
\hline
\cellcolor{gray!15} \textbf{model} & \cellcolor{gray!15} \textbf{weight dec.} & \cellcolor{gray!15} \textbf{patience} & \cellcolor{gray!15} \textbf{batch Size} & \cellcolor{gray!15} \textbf{dropout} \\ \hline
SAGPool  & 0.0001 & 100 & 128 & 50\% \\ \hline
GMT & 0.0001 & 50 & 128 & 50\% \\ \hline
DiffPool & Default & 50 & 128 & No \\ \hline
DMon & Default & 50 & 128 & No \\ \hline
MinCut & 0.0001 & 50 & 128 & No \\ \hline
AdamGNN & Default & 50 & 64 & 50\% \\ \hline
HGP-SL & 0.001 & 50 & 512 & No \\ \hline
GIN-($\epsilon$ \& $0$) & 0.5 & No & 128 & 50\% \\ \hline
GCN & 0.0001 & 100 & 128 & 50\% \\ \hline
\end{tabular}
\label{table:param_details_2}
\end{table}

\subsection{Parameter Details}
Table~\ref{table:param_details_1} and~\ref{table:param_details_2} represent all the baseline models' parameters' details. A notable observation is that the number of maximum epochs for the SAGPool and MinCutPool seems endless. However, due to the patience variable, models take a much smaller number of epochs during the experiment. The AdamGNN model determines the number of layers for the experiment by analyzing the graphs' structural properties. All the models are developed in the PyTorch library and utilize the Adam optimizer where the default weight decay is set to $0$ in most cases. Except for AdamGNN (64) and HGP-SL (512), the batch size is 128 for all of the other models. All the baseline models employ different learning rates for evaluation. Six of the models employ the dropout parameter with a rate of 50\%, while the other four models abstain from utilizing it. 

% \appendices
% \section{Appendix A}
% Your appendix content goes here.

\end{document}